\documentclass{egpubl}
\usepackage{pg2020}
 
\SpecialIssuePaper         

\usepackage[T1]{fontenc}
\usepackage{dfadobe}  

\biberVersion
\BibtexOrBiblatex
\usepackage[backend=biber,bibstyle=EG,citestyle=alphabetic,backref=true]{biblatex} 
\AtEveryBibitem{%
\ifentrytype{article}{%
}{}
\ifentrytype{inproceedings}{%
  \clearfield{volume}
  \clearfield{number}
}{}
  \clearlist{publisher}
  \clearfield{isbn}
  \clearlist{location}
  \clearfield{doi}
  \clearfield{url}
}
\electronicVersion
\PrintedOrElectronic

\ifpdf \usepackage[pdftex]{graphicx} \pdfcompresslevel=9
\else \usepackage[dvips]{graphicx} \fi

\usepackage{egweblnk} 

\usepackage{amsmath}
\usepackage{amsfonts}
\usepackage{egweblnk} 
\usepackage{breqn}
\usepackage{color}

\newcommand{\revA}[1]{#1}
\newcommand{\revB}[1]{#1}
\newcommand{\revC}[1]{#1}
\newcommand{\revD}[1]{#1}
\newcommand{\revE}[1]{#1}
\newcommand{\revF}[1]{#1}
\newcommand{\revG}[1]{#1}
\newcommand{\revH}[1]{#1}
\newcommand{\revI}[1]{#1}
\newcommand{\revJ}[1]{#1}
\newcommand{\chkURL}[1]{\textcolor{magenta}{#1}}

\title[Diversifying Semantic Image Synthesis and Editing via \revD{Class-} \revC{and Layer-wise VAEs}]%
      {Diversifying Semantic Image Synthesis and Editing \\via \revD{Class-} \revC{and Layer-wise VAEs}}

\author[Y. Endo \& Y. Kanamori]
{\parbox{\textwidth}{\centering Y. Endo\orcid{0000-0001-5132-3350}
        and Y. Kanamori\orcid{0000-0003-2843-1729} 
        }
        \\
{\parbox{\textwidth}{\centering University of Tsukuba, Japan
       }
}
}

\begin{document}

\teaser{
 \includegraphics[width=\linewidth]{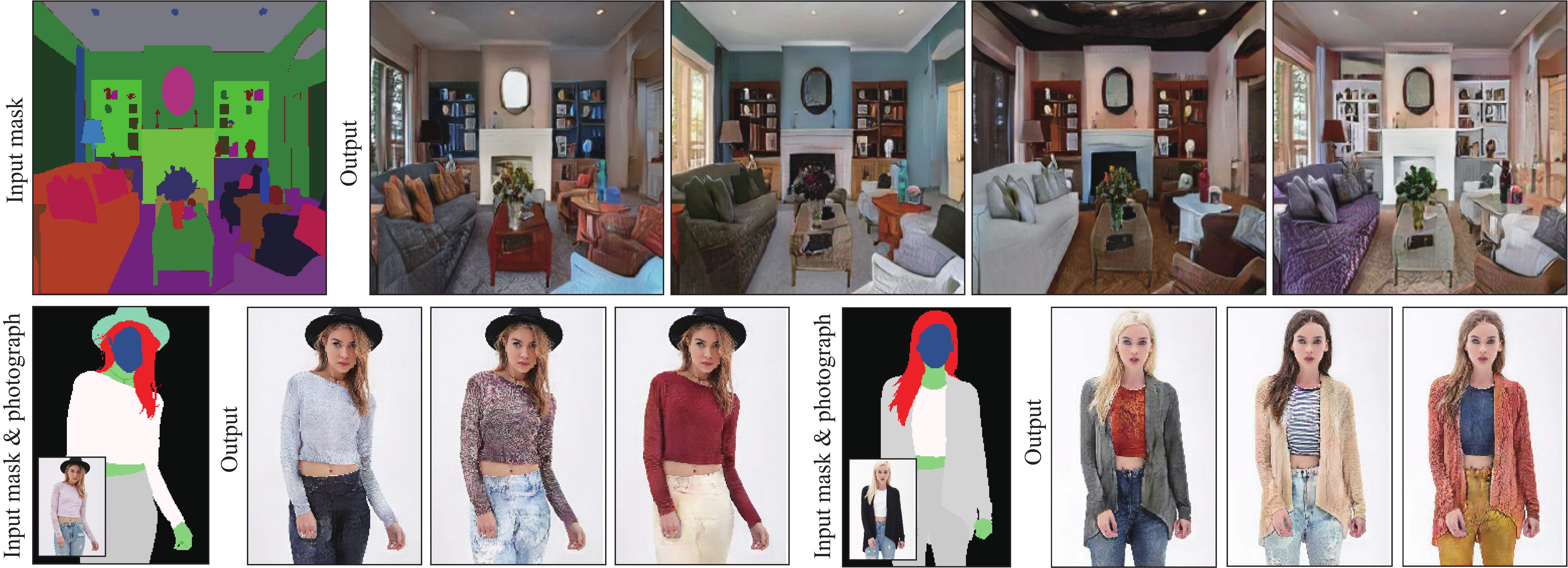}
 \centering
  \caption{
    \revC{Results of} multimodal semantic image synthesis and editing \revC{using} \revD{our} \revC{method}.
    Our method \revA{yields highly} diverse images from a single semantic mask (top)\revA{, and also} enables \revA{appearance editing for specific semantic objects, e.g., the clothes in the fashion} \revJ{images} (bottom). 
  }
\label{fig:teaser}
}

\maketitle
\begin{abstract}
\revA{Semantic image synthesis} \revC{is a process for generating} photorealistic images from a single semantic mask. 
\revA{\revC{To enrich} the diversity \revC{of} multimodal image synthesis, previous methods \revC{have controlled}} the global appearance of an output image by learning a single latent space. 
However, \revA{a single latent code is often insufficient} \revC{for capturing} various object styles because object appearance depends on \revC{multiple} factors. 
\revA{To handle} individual factors that determine object styles, we propose a class- and layer-wise \revA{extension to the \textit{variational autoencoder} (VAE) framework}
\revC{that} allows \revA{flexible control over} each object class at \revC{the} \revA{local to global} levels by learning multiple latent spaces. 
\revA{\revC{Furthermore, w}e demonstrate that our method generates \revC{images that are both plausible and more diverse} compared to state-of-the-art methods via} extensive experiments with real and \revA{synthetic} datasets \revJ{in} three different domains.
We also \revA{show} that our method \revA{enables} a wide range of \revA{applications in} image synthesis and editing tasks. Codes are available at~\href{https://github.com/endo-yuki-t/DiversifyingSMIS}{\chkURL{https://github.com/endo-yuki-t/DiversifyingSMIS}}. 


\end{abstract}  
\section{Introduction}\label{sec:intro}

\revA{Semantic image synthesis is a fascinating technique \revC{for converting} a semantic mask into natural images. 
This technique has been gaining considerable attention because of its broad applications such as design exploration and content creation.
Ideally, for these practical applications, semantic image synthesis should generate realistic \revC{and} diverse images from a single semantic mask. 
\revC{Thus} we seek \textit{multimodal} image synthesis, which entails a one-to-many mapping problem.}

\revA{Semantic image synthesis has advanced \revC{significantly} since the emergence of \textit{conditional generative adversarial \revC{networks}} (\revC{cGANs})~\cite{DBLP:conf/cvpr/IsolaZZE17}. 
\revC{Early cGAN techniques were limited to \textit{unimodal image synthesis} (i.e., one-to-one mapping from a semantic mask to an output image) \cite{DBLP:conf/cvpr/Wang0ZTKC18}. 
\revJ{Multimodal} image synthesis uses several approaches, such as the \textit{variational autoencoder} (VAE)~\cite{DBLP:journals/corr/KingmaW13}, combinations of VAE and GAN (VAE-GAN)~\cite{DBLP:conf/icml/LarsenSLW16,DBLP:conf/nips/ZhuZPDEWS17,DBLP:conf/cvpr/Park0WZ19,liu2019learning}, and \textit{implicit maximum likelihood estimation} (IMLE)~\cite{DBLP:journals/corr/abs-1809-09087,DBLP:conf/iccv/LiZM19}.}
}
\revA{\revC{Nonetheless}, the variety of output images is still limited because a single latent code controls the entire content in each image.
\revC{Consider the} interior design of indoor scenes (\revD{e.g.,} the top row of Figure~\ref{fig:teaser}). 
\revC{We should be able to change the design of the} furniture (e.g., a sofa, desk, and shelf) according to \revE{users'} \revC{preferences}.} 
A single latent representation is \revA{thus} insufficient to capture such style variations adequately.

\revA{In this \revC{study}, we enhance the VAE-GAN framework to introduce flexible control over each semantic class while supporting a large number of class labels. 
Unlike previous \revC{approaches} where a single latent code is extracted only from the last layer of the encoder, our \revD{key idea} is to learn the class-wise latent space from each layer of the encoder \revC{to capture} multiple factors that affect the class-wise object appearance. 
\revC{Next,} class-wise latent codes are extracted from the intermediate layers of the encoder using the semantic mask, spatially replicated according to the mask's spatial structure, and then fed to the corresponding generator blocks.
\revC{Hence,} we can synthesize \revC{the} diverse image content for each class. 
\revC{The} layer-wise latent \revD{codes allow} flexible image editing at local (e.g., textures) to global (e.g., entire color tones) levels, as \revC{proposed} in \textit{StyleGAN}~\cite{DBLP:conf/cvpr/KarrasLA19,DBLP:journals/corr/abs-1912-04958}.
The linear functions for sampling \revC{the} latent codes are shared among the  class-wise latent codes in each layer.
Therefore our approach is independent of and thus scalable \revC{with} the number of class labels. 
To suppress \revC{the} degradation of generalization performance, we introduce additional regularization for latent codes during training.}

Our class- and layer-wise \revE{VAEs are} a simple yet \revA{powerful} extension for modeling various styles more efficiently than state-of-the-art methods. 
\revA{We demonstrate that our method can generate \revC{images that are both plausible and more diverse} than \revC{the} competing methods via experiments with} \revC{the following} datasets\revC{:} \textit{ADE20K}~\cite{DBLP:conf/cvpr/ZhouZPFB017}, \textit{DeepFashion}~\cite{liuLQWTcvpr16DeepFashion}, and \textit{GTA-5}~\cite{DBLP:conf/eccv/RichterVRK16}. 
\revA{\revC{In addition,} we show that our method has interesting applications in image synthesis \revC{and} editing tasks\revC{,} such as \revC{the} exploration of interior or \revC{clothing} design, as illustrated in Figure~\ref{fig:teaser}.}  

\section{Related Work}
\subsection{Conventional Image Editing}

\revA{Various} image editing techniques have been proposed \revA{in the field of computer graphics.}
Image analogies~\cite{DBLP:conf/siggraph/HertzmannJOCS01} can apply a transformation between a pair of aligned images to another image.
\revC{Certain} methods use reference images in a database to edit objects and textures in a target image~\cite{DBLP:journals/tog/ChenCTSH09, DBLP:journals/tog/HaysE07,DBLP:journals/tog/LalondeHERWC07}.
These non-parametric methods can produce photorealistic results if \revC{the} reference images \revC{retrieved from the database match target images}. 
However, the generalization performance for various inputs is low, and the cost of \revC{running} the database is high. 

\subsection{Generative Adversarial Networks}

\revA{A} parametric approach for image generation has evolved \revJ{since} the advent of \textit{generative adversarial networks} (GANs)~\cite{DBLP:conf/nips/GoodfellowPMXWOCB14}, \revA{and} the quality of the generated images has greatly improved.
The recent GAN-based technique known as \textit{StyleGAN}~\cite{DBLP:conf/cvpr/KarrasLA19,DBLP:journals/corr/abs-1912-04958} can produce high quality and diverse images via feature normalization using \textit{adaptive instance normalization} (AdaIN)~\cite{DBLP:conf/iccv/HuangB17} and latent \revA{code} regularization. 
\revA{While} GANs \revC{were} originally \revA{designed} to generate \revA{various outputs from random noise}, they have been \revC{adapted} to a variety of tasks by conditioning \revA{them} on \revA{input data, i.e., \textit{conditional GANs} (cGANs)}.

\subsection{Image-to-Image Translation}

\revA{A successful example of cGANs is the image-to-image translation, \revC{in which} an input image is converted to another image domain.}
\revD{\textit{Pix2pix}}~\cite{DBLP:conf/cvpr/IsolaZZE17} is \revA{a} seminal work of image-to-image translation \revA{that showcased} image domains such as sketches, maps, and semantic masks. 
\revC{Moreover, }\revA{pix2pix was extended to support higher resolution images~\cite{DBLP:conf/cvpr/Wang0ZTKC18} and unsupervised settings~\cite{DBLP:conf/iccv/ZhuPIE17,DBLP:conf/cvpr/ChoiCKH0C18}.}
The encoder-decoder architectures for image-to-image translation also play \revA{an} important role in various specific tasks in computer graphics~\cite{DBLP:journals/tog/PortenierHSBFZ18,DBLP:journals/tog/BauSPWZZ019,DBLP:journals/tog/EilertsenKDMU17,DBLP:journals/tog/EndoKM17,DBLP:journals/tog/ZhangLW0L18,DBLP:journals/tog/KanamoriE18,DBLP:journals/tog/SunBTXYFRBDR19}. 

\subsection{Semantic Image Synthesis}

\revA{Synthesizing realistic and diverse images from a semantic mask is an ill-posed and challenging problem, which we \revC{address} in this \revC{study}.}
\revA{Although pix2pix supports semantic image synthesis, it suffers from \textit{mode collapse}; \revC{that is,} pix2pix obtains only a single plausible image even \revC{when using} latent noise as additional input to control the output.}
To \revC{solve} this problem, various methods for multimodal semantic image synthesis have been proposed.
\revD{The} cascaded refinement network (CRN)~\cite{DBLP:conf/iccv/ChenK17} \revA{outputs a fixed number of images while encouraging their diversity via \revJ{the} diversity loss in the training phase.}
Ghosh et al. achieved multimodal image \revA{synthesis with} multiple generators \revA{trained for} different modes~\cite{DBLP:conf/cvpr/GhoshKNTD18}. 
However, these methods can only generate a fixed number of images from a single input image.

To handle a variable number of modes while avoiding mode collapse, 
\revA{the} variational autoencoder (VAE)~\cite{DBLP:journals/corr/KingmaW13} is used \revC{jointly} with \revA{the} GAN\revC{; this} is also known as \revA{the} VAE-GAN~\cite{DBLP:conf/icml/LarsenSLW16}. 
\revD{BicycleGAN}~\cite{DBLP:conf/nips/ZhuZPDEWS17} \revA{employed a conditional version of the VAE-GAN with a latent regressor~\cite{DBLP:conf/iclr/DonahueKD17,DBLP:conf/iclr/DonahueKD17,DBLP:conf/nips/ChenCDHSSA16} for multimodal image synthesis}. 
Most recently, Li et al.~\cite{DBLP:conf/iccv/LiZM19} \revA{used} \textit{conditional implicit maximum likelihood estimation} (cIMLE)~\cite{DBLP:journals/corr/abs-1809-09087} instead of GANs. 
Unlike BicycleGAN\revA{, which uses multiple networks such as generators, discriminators, and encoders, they stabilized the training using a single network.} 
\textit{GauGAN}~\cite{DBLP:conf/cvpr/Park0WZ19} improved the quality of semantic image synthesis \revA{significantly} using normalization suitable for semantic masks.
\revA{Its VAE-GAN variant} \revC{facilitates} high-quality multimodal image synthesis.
\revA{Subsequent studies \revC{have} also improved} the quality \revC{of image synthesis~\cite{liu2019learning}}.
However, all these methods use a single latent code for a single input mask. 
\revC{By} contrast, our method enables diverse image synthesis and flexible editing by learning \revC{the} latent spaces per class and \revC{per} layer. 

A \revA{concurrent} study~\cite{conf/cvpr/zhu20} \revA{diversifies multimodal image synthesis with} a similar motivation to our work\revC{; namely,} leveraging group convolutions.
Their \revC{method showed} good performance for datasets \revA{with a small number of \revA{semantic} classes (e.g., eight in a simplified version of \textit{DeepFashion})}, but it \revC{struggled} to handle a \revA{large} number of classes.
\revA{\revC{By} contrast, our} method \revA{is scalable \revC{regarding} the number of classes (see Section~\ref{sec:intro}) and \revC{significantly} outperforms \revJ{the} \revC{aforementioned} method in datasets with many classes (see Section~\ref{sec:experiments})}. 

\section{Method}

Our method \revA{adopts the VAE-GAN framework}. 
For the generator, we employ \revA{a} modified version of GauGAN~\cite{DBLP:conf/cvpr/Park0WZ19}, which is a state-of-the-art method capable of generating high-quality images from a semantic mask.
In this section, we first briefly review GauGAN and then explain our class- and layer-wise \revE{VAEs}. 

\subsection{Background}

\revD{GauGAN}~\cite{DBLP:conf/cvpr/Park0WZ19} \revA{introduces} a normalization module called \textit{spatially adaptive denormalization} (SPADE).
\revC{The SPADE module} normalizes an input feature map from the previous layer and then updates the feature map using modulation parameters, which are calculated \revC{using} convolutions for the input semantic mask. 
Unlike other normalization layers, the modulation parameters have spatially different values and \revA{are} used via pixel-wise multiplication and summation.
\revC{Thus,} SPADE successfully retains the spatial information of the semantic mask in the feature map. 
We \revC{have omitted} the details of \revC{the SPADE module} because \revC{they are} not relevant to our method.

\begin{figure}[t]
   \centering
   \includegraphics*[width=1.\linewidth, clip]{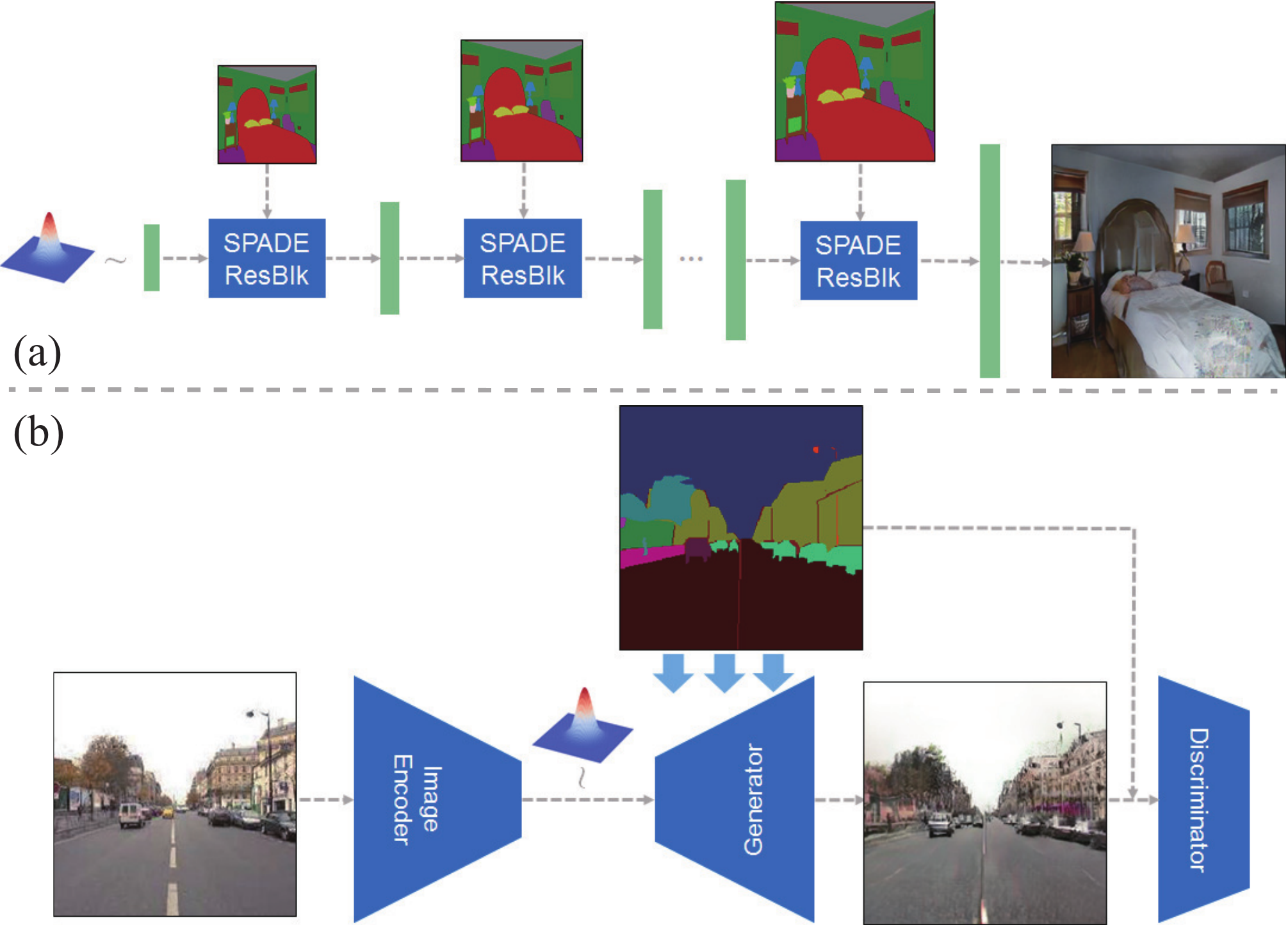}
   \caption{
\revC{Network} architecture of the SPADE generator (a) and the training pipeline (b).
}
   \label{fig:spade}
\end{figure}

Figure~\ref{fig:spade} \revC{(a)} illustrates the SPADE generator, which is a lightweight network consisting of a decoder only. 
Given \revC{the} semantic mask $\mathbf{M} \in \{0, 1\}^{C \times W \times H}$ (\revA{where} $C$, $W$, and $H$ are the total number of classes, width, and height) \revA{in a one-hot representation}, the SPADE generator $G$ calculates an output image as follows: 
\begin{align}
 G(\mathbf{M})  = g_{K}(\revA{\mathbf{M}_K}, g_{K-1}(\revA{\mathbf{M}_{K-1}}, \cdots g_{1}(\revA{\mathbf{M}_1}))),
\end{align}
where $g_k$ is a $k$th block, and $K$ is the number of blocks.
\revA{$\mathbf{M}_k \in \{0,1\}^{C \times W_k \times H_k}$ is a downsampled semantic mask whose width $W_k$ and height $H_k$ are the same as those of the corresponding feature map.}
In the case of unimodal image synthesis, the first block $g_1$ is a simple \revC{convolutional} layer that \revA{takes} $\mathbf{M}_1$ as \revA{an} input. Meanwhile, $g_k$ ($k \geq 2$) consists of the SPADE ResNet block, which \revA{takes} an intermediate feature map as well as $\mathbf{M}_k$ as \revA{inputs}. 

In the case of multimodal image synthesis based on GauGAN-VAE~\cite{DBLP:conf/cvpr/Park0WZ19}, 
a latent code $\mathbf{z}_1 \in \mathbb{R}^{N_1}$ \revA{is used} as an additional input \revC{as follows}: 
\begin{align}
 G(\mathbf{M}, \mathbf{z}_1)  = g_{K}(\revA{\mathbf{M}_K}, g_{K-1}(\revA{\mathbf{M}_{K-1}}, \cdots g_{1}(\mathbf{z}_1))).
 \end{align}
In this case, $g_1$ consists of linear and \revA{reshaping} functions\revC{, which} \revJ{acquires} a feature map from the input vector. 
The latent code $\mathbf{z}_1$ is sampled from the prior at \revJ{the} test time: 
\begin{align}
\mathbf{z}_1 \sim \mathcal{N}(\mathbf{0},I), 
\end{align}
\revG{where $I$ is an indentity matrix. }

At the training time shown in Figure~\ref{fig:spade} \revC{(b)}, given a ground-truth RGB image~$\mathbf{I} \in \mathbf{R}^{3 \times W \times H}$, the latent code $\mathbf{z}_1$ is sampled from the encoder $E$ via the reparameterization trick~\cite{DBLP:journals/corr/KingmaW13}:
\begin{align}
 \mathbf{z}_1 &\sim E(\mathbf{I}) = \mathcal{N}(\mu(\mathbf{I}), \sigma^2(\mathbf{I})), \\
\mu(\mathbf{I}) &= f^{\mu}(e_{L}(e_{L-1}(\cdots e_{1}(\mathbf{I})))), \\
\sigma^2(\mathbf{I}) &= f^{\sigma^2}(e_{L}(e_{L-1}(\cdots e_{1}(\mathbf{I})))), 
\end{align}
where $L$ denotes the number of encoder blocks, and $e_l$ is the $l$th block \revC{comprising the convolutional}, instance normalization, and LeakyReLU layers. 
The functions $f^{\mu}$ and $f^{\sigma^2}$ are \revA{linear} layers and compute the mean and the variance from the feature map. 

To train the generator $G$ and the encoder $E$, we minimize the following loss function: 
\begin{align}
  \revJ{\mathcal{L}(G, D, E)}  = \; & \mathcal{L}_{GAN}(G, D, E) + \lambda_{FM} \mathcal{L}_{FM}(G, E)\nonumber \\ &+ \lambda_{VGG} \mathcal{L}_{VGG}(G, E) + \lambda_{KL} \mathcal{L}_{KL}(E),  \label{eq:loss2} \\
  G^*, E^*  = \; & \arg \min_{G,E} \max_{D} \revJ{\mathcal{L}(G, D, E)},&
\end{align}
where $D$ is the discriminator\revC{;} $\mathcal{L}_{GAN}$, $\mathcal{L}_{FM}$, and $\mathcal{L}_{VGG}$ are adversarial~\cite{DBLP:conf/nips/GoodfellowPMXWOCB14}, GAN feature matching~\cite{DBLP:conf/cvpr/Wang0ZTKC18}, and perceptual \revC{losses}~\cite{DBLP:conf/eccv/JohnsonAF16}, respectively. 
\revC{In addition}, $\mathcal{L}_{KL}$ is the regularization term for the VAE: 
\begin{align}
\mathcal{L}_{KL}(E) = \mathbb{E}_{\mathbf{I} \sim p(\mathbf{I})}[\mathcal{D}_{KL}(E(\mathbf{I})\|\mathcal{N}(0,I))], 
\end{align}
where $\mathcal{D}_{KL}$ is \revA{the} Kullback-Leibler divergence (KLD), and $\lambda_{FM}$, $\lambda_{VGG}$, and $\lambda_{KL}$ are \revC{the} empirically determined weights.

\subsection{Class- and Layer-wise \revE{VAEs}}

\begin{figure}[t]
   \centering
   \includegraphics*[width=1.\linewidth, clip]{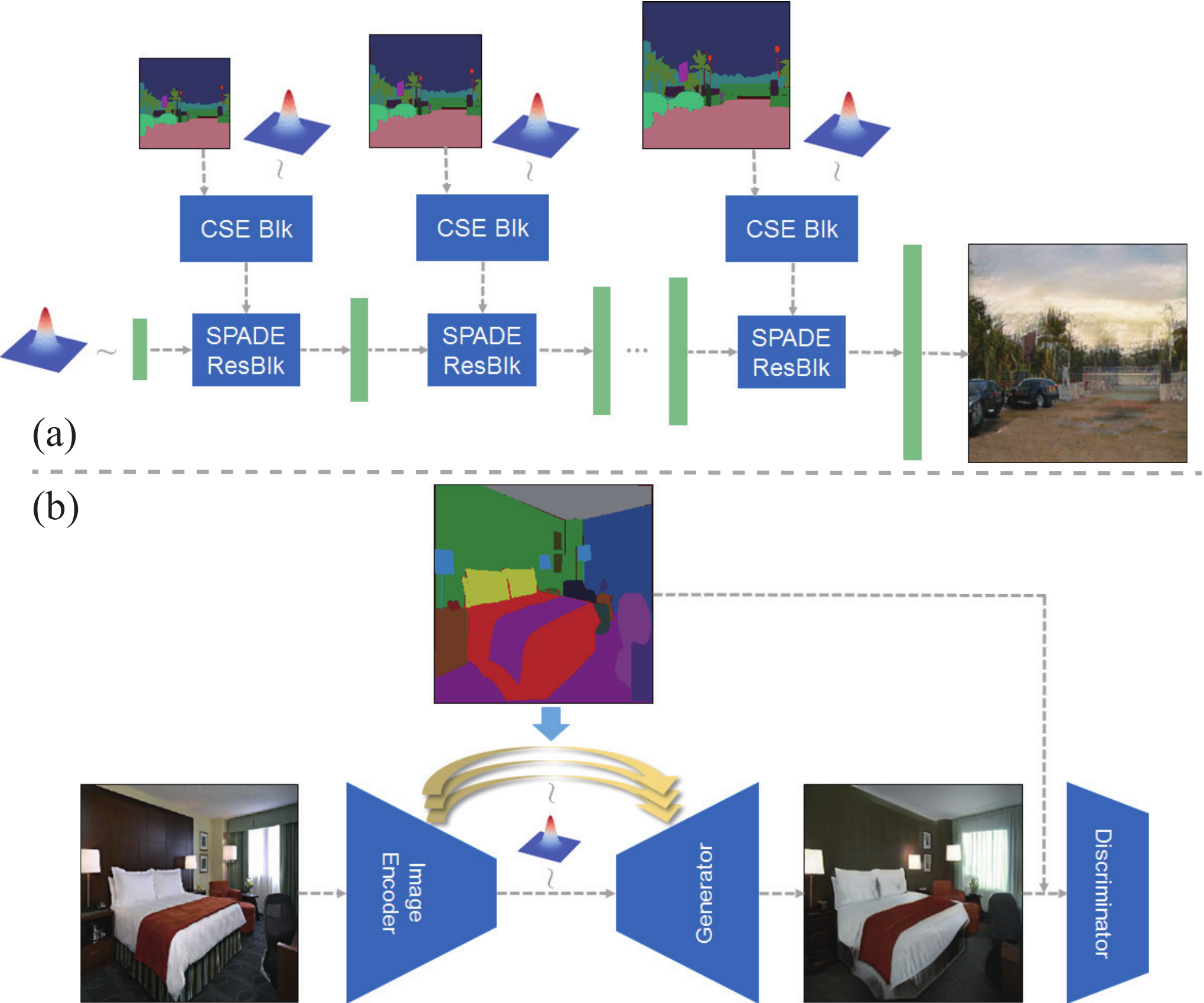}
   \caption{
\revC{Network} architecture of our generator (a) and the training pipeline (b).
}
   \label{fig:overview}
\end{figure}

\revA{\revC{In addition to} the VAE-GAN framework explained \revC{earlier}, we introduce \revC{the} class- and layer-wise style controls} using multiple latent codes. 
Figure~\ref{fig:overview} illustrates our generator and the \revC{entire} training pipeline. 
\revA{During training}, \revA{our encoder extracts class-wise} latent codes from the feature \revA{maps} obtained from the final layer \revC{and} from each \revC{individual} layer. 
Our generator receives latent codes at the first block \revC{and} at each \revC{subsequent} block, as shown in the bottom of Figure~\ref{fig:overview}.
\revA{Our} \textit{class-wise semantics embedding} (CSE) block (Figure~\ref{fig:block}) plays \revA{the key} role in \revA{exploiting \revC{the} class-wise} latent codes \revA{at the generator blocks}. 
We explain the generator and the encoder \revC{in the following sections}. 

\subsubsection{Generator with CSE blocks}

First, we define a set of latent codes fed to the generator as $\mathcal{Z} = \{\mathbf{z}_1, \mathbf{Z}_{2}, \mathbf{Z}_{3}\, \cdots, \mathbf{Z}_{L}\}$, where $L \leq K$. 
Here, $\mathbf{z}_1 \in \mathbb{R}^{N_1}$ \revA{(where $N_1$ is the number of $\mathbf{z}_1$ channels)} is a vector that controls the entire appearance of an output image\revA{, similar} to \revC{the} GauGAN-VAE~\cite{DBLP:conf/cvpr/Park0WZ19}.
$\mathbf{Z}_l = (\mathbf{z}_{l, 1}, \mathbf{z}_{l, 2}, \cdots, \mathbf{z}_{l, C}) \in \mathbb{R}^{N\times C}$ \revA{(where $N$ is also the number of channels)} is a 2D tensor consisting of latent codes for $C$ classes \revA{and is fed to \revC{the} generator} block~$l$ \revA{corresponding to \revC{the} encoder's \revB{block}~$l$}.
The generator $G$ gives an output image from a semantic mask $\mathbf{M}$ and a set of latent codes $\mathcal{Z}$:
\begin{dmath}
 G(\mathbf{M}, \mathcal{Z}) \! = \! g_K\!(s(\revA{\mathbf{M}_{L}}, \mathbf{Z}_{L}), g_{K\!-\!1}\!(s(\revA{\mathbf{M}_{L\!-\!1}}, \mathbf{Z}_{L\!-\!1}), \cdots g_{1}\!(\mathbf{z}_1))), \label{eq:G}
\end{dmath}
where $s$ is the CSE block, which generates a feature map by combining a \revA{downsampled} semantic mask \revA{$\mathbf{M}_l$ and \revC{the} class-wise} latent codes \revA{$\mathbf{Z}_l$}.
The feature maps \revA{computed in $s$} are fed to the SPADE ResNet blocks. 
\revA{\revC{Similar} to $\mathbf{z}_1$, $\{\mathbf{z}_{l,c}\}_{c \in \{1, 2, \dots, C\}}$} are sampled from the standard \revA{normal} distribution at \revA{the} test time: 
\begin{align}
  \mathbf{z}_{l ,c} \sim \mathcal{N}(\mathbf{0},I). 
\end{align}
\revA{In \revC{cases where} the number of generator blocks $K$ is larger than the number of encoder \revB{blocks} $L$, latent codes for \revC{the} $(K-L)$ generator blocks are missing.
\revJ{This holds true for our experiments because we used the same number of blocks as GauGAN (i.e., $L<K$) for fair comparison.}
In this case, we \revC{suggest} two options\revC{:} simply \revG{replacing} $s$ with $\mathbf{M}_l$ or \revG{reusing} the same latent code at \revC{the} $(K-L)$ blocks.}
\revJ{In our experiments,} we adopted \revA{a combination of} these two options\revA{, as shown in the supplementary material}.
\begin{figure}[t]
   \centering
   \includegraphics*[width=1.\linewidth, clip]{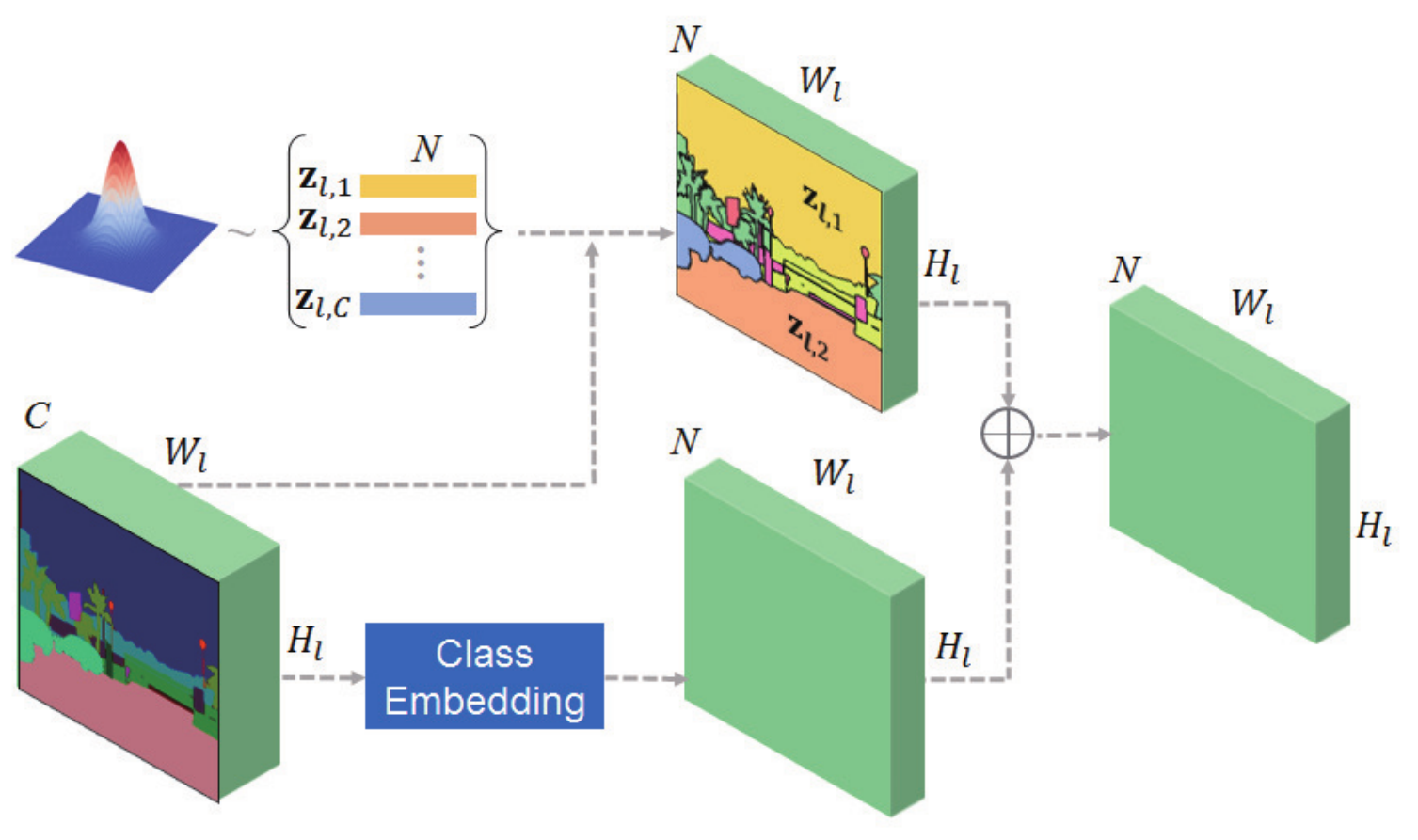}
   \caption{
\revA{\revC{Class-wise} semantics embedding (CSE)} block. \revA{In the upper path, we replace the one-hot vectors in the semantic mask with the corresponding latent codes. In the lower path, we embed the semantic mask in the same dimension as \revJ{the counterpart in the upper path}. The final feature map is obtained by their element-wise addition.}
}
   \label{fig:block}
\end{figure}

\revA{Figure~\ref{fig:block} shows the} design of the CSE block.
The feature map extracted by the CSE block is \revC{given as follows}: 
\begin{align}
 s(\revA{\mathbf{M}_l}, \mathbf{Z}_l) =  \mathcal{T}^{-1}\!(\mathbf{Z}_l \, \mathcal{T}\!(\revA{\mathbf{M}_l}) + \mathbf{W} \, \mathcal{T}\!(\revA{\mathbf{M}_l})), 
\end{align}
where $\mathcal{T}$ is a \revA{reshaping} function\revA{, i.e.,} $\mathcal{T}: \mathbb{R}^{C \times \revA{W_l} \times \revA{H_l}} \rightarrow \mathbb{R}^{C \times \revA{W_l \, H_l}}$. 
In the first term (the \revA{upper part} of Figure~\ref{fig:block}), 
\revA{we replace \revC{the} class-wise one-hot vectors with the corresponding class-wise latent codes.}
\revA{Note that the latent codes in the first term are random vectors and thus do not \revC{contain} the \revJ{objects' class} information (e.g., chair and floor) that they represent.}
In the second term (the \revA{lower part} of Figure~\ref{fig:block}), \revA{to balance with the first term's dense representation,} we embed the sparse \revA{(i.e., one-hot)} vectors \revA{in} the semantic mask into dense vectors of the same dimension as the latent code \revA{by multiplying \revC{the} trainable} weight $\mathbf{W} \in \mathbb{R}^{N \times C}$. 
\revA{\revC{Exploiting} the sparsity of $\mathbf{M}_l$'s one-hot representation, we can \revC{quickly} calculate \revC{the} matrix multiplications in the first and second terms.}
We can obtain \revA{the} final feature map via element-wise summation of these two feature maps.

\subsubsection{Encoder for learning latent codes}

\begin{figure}[t]
   \centering
   \includegraphics*[width=1.\linewidth, clip]{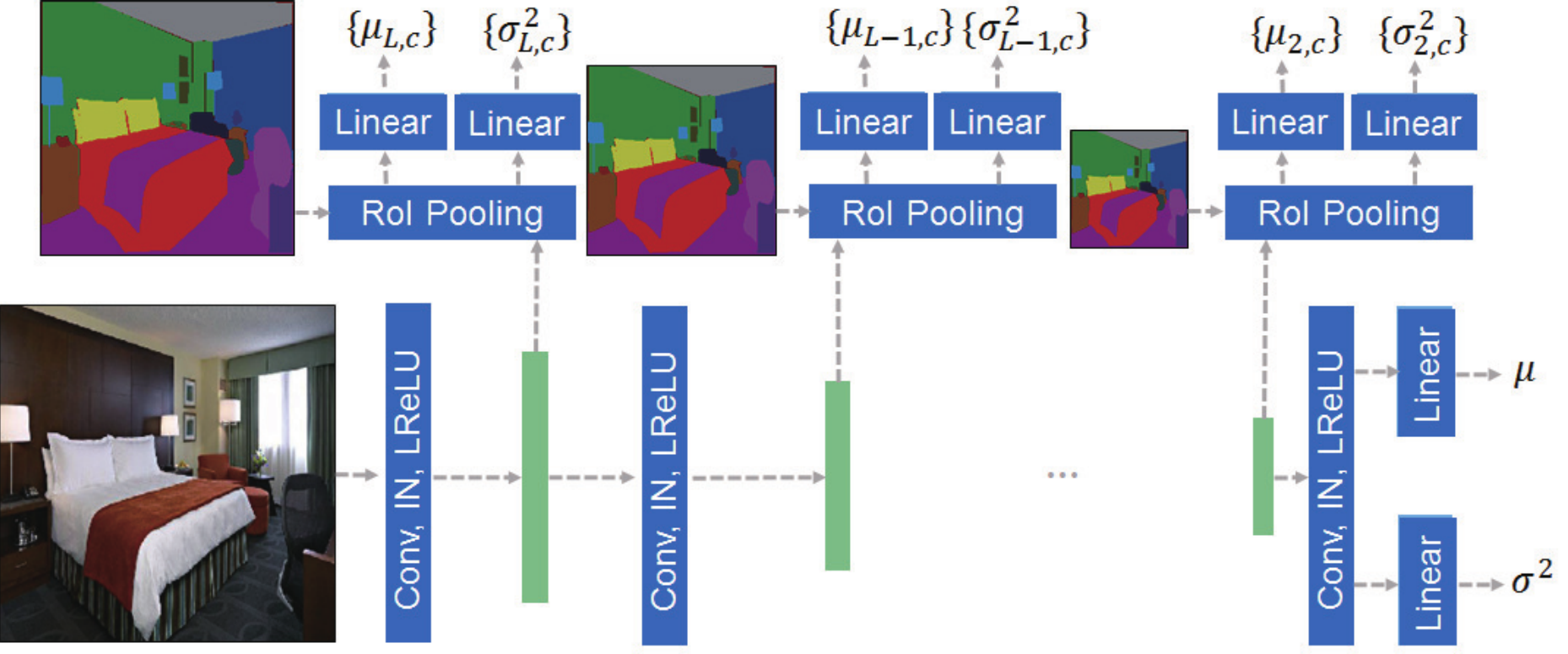}
   \caption{
\revC{Encoder} architecture. 
\revC{The} means and variances of \revC{the} latent code distributions \revC{are calculated} from a feature map \revA{at} each \revB{block}.
The RoI pooling allows \revA{us to extract} vectors \revC{with} the same \revC{dimension} from regions \revC{of different sizes}.}
   \label{fig:encoder}
\end{figure}

\revA{\revC{In this section,}} we describe the encoder for extracting latent codes for each class and layer.
Figure~\ref{fig:encoder} illustrates our encoder design. 
\revA{Recall that \revC{the GauGAN-VAE} encoder} samples \revC{the} single latent code \revA{$\mathbf{z}_1$ dependent} on the entire feature \revA{map at} the final layer. 
In addition to that, our method samples a latent code $\mathbf{z}_{l,c}$ for \revA{each} class $c$ and \revA{each} \revB{block} $l$:
\begin{align}
 \mathbf{z}_{l,c} \sim E_{l,c}(\mathbf{I}) = \mathcal{N}(\mu_{l,c}(\mathbf{I}), \sigma^2_{l,c}(\mathbf{I})). 
\end{align}
We compute the mean and the variance from \revC{the feature map of each block} on a region \revA{$\Omega_{l,c}$} \revA{that contains \revC{the} class label $c$ in the semantic mask $\mathbf{M}_l$}:
\begin{align}
 \mu_{l, c}(\mathbf{I}) &= f^\mu_l(r(\revA{\Omega_{l,c}}, e_{L-l+1}(e_{L-l}(\cdots e_1(\mathbf{I})))), \\
 \sigma^2_{l, c}(\mathbf{I}) &= f^{\sigma^2}_l(r(\revA{\Omega_{l,c}}, e_{L-l+1}(e_{L-l}(\cdots e_1(\mathbf{I})))), 
\end{align}
where $f^\mu_l$ and $f^{\sigma^2}_l$ are \revC{the} \revA{linear} \revJ{functions} for \revB{\revC{the}} block $l$.
\revA{These linear \revJ{functions} are shared among \revC{the} classes $c$ in each \revB{block} $l$ when we sample \revC{the} latent codes $\mathbf{z}_{l,c}$, as explained in Section~\ref{sec:intro}}. 
Although the number of pixels in \revC{the} region \revA{$\Omega_{l,c}$} differs depending on class $c$, \revA{the extracted feature \revC{vectors} for each class should have the same \revC{dimension}}. 
To \revC{manage} this, we use a \revA{region-of-interest} (RoI) pooling $r$ that extracts a fixed-sized vector from a feature map~\revA{$\mathbf{H}$} (3D tensor): 
\begin{align}
 r (\revA{\Omega_{l,c}}, \revA{\mathbf{H}}) &= (H'_{1, c}, \, H'_{2, c}, \, \cdots, \, \revA{H'_{j, c}}, \, \revA{\cdots,} \, H'_{n, c})^T, \\
 \revA{H'_{j, c}} &= \revA{\max_{(x, y) \in \Omega_{l,c}} H_{j, x, y}},
\end{align}
where $n$\revA{, $x$, and $y$ are the number of output channels, $x$ \revC{coordinate} and $y$ \revC{coordinate} in \revC{the} region $\Omega_{l,c}$}. 

For training the networks, we use \revA{a} modified version of Equation~(\ref{eq:loss2}). 
Specifically, we redefine the regularization term $\mathcal{L}_{KL}$ to handle multiple outputs \revC{from} our encoder: 
\begin{align}
\mathcal{L}_{KL}(E) = \; & \mathbb{E}_{\mathbf{I}, \mathbf{M} \sim p(\mathbf{I},\mathbf{M})}[\mathcal{D}_{KL}(E(\mathbf{I})\|\mathcal{N}(0,I))\nonumber \\ 
&+ \frac{1}{|\mathcal{C}(\mathbf{M})|}\Sigma_{l=2}^{L} \Sigma_{c \in \mathcal{C}(\mathbf{M})} \mathcal{D}_{KL}(E_{l,c}(\mathbf{I})\|\mathcal{N}(0,I))], 
\end{align}
where $\mathcal{C}(\mathbf{M}$) is a function that returns a set of unique classes existing in \revC{the} semantic mask $\mathbf{M}$. 
The first term is for a latent code $\mathbf{z}_1$ \revA{\revC{that is} dependent} on \revA{the} global appearance of \revA{the} input image\revC{;} the second term is for latent codes \revA{$\{\mathbf{Z}_{l,c}\}$ \revC{that are} dependent} on each class and \revA{each} \revB{block}. 
To prevent the regularization from being biased toward images containing many unique classes, we normalize the sum of \revC{the} KLD with the number of \revA{unique} classes. 
For the \revA{hyper-parameters}, we used the same values as in \revD{the GauGAN-VAE}~\cite{DBLP:conf/cvpr/Park0WZ19}\revC{; specifically,} $\lambda_{FM}=10$, $\lambda_{VGG}=10$ and $\lambda_{KL}=0.05$. 
The details of our network architectures \revA{can be found} in the supplementary materials.

\begin{figure*}[t]
   \centering
   \includegraphics*[width=1.\linewidth, clip]{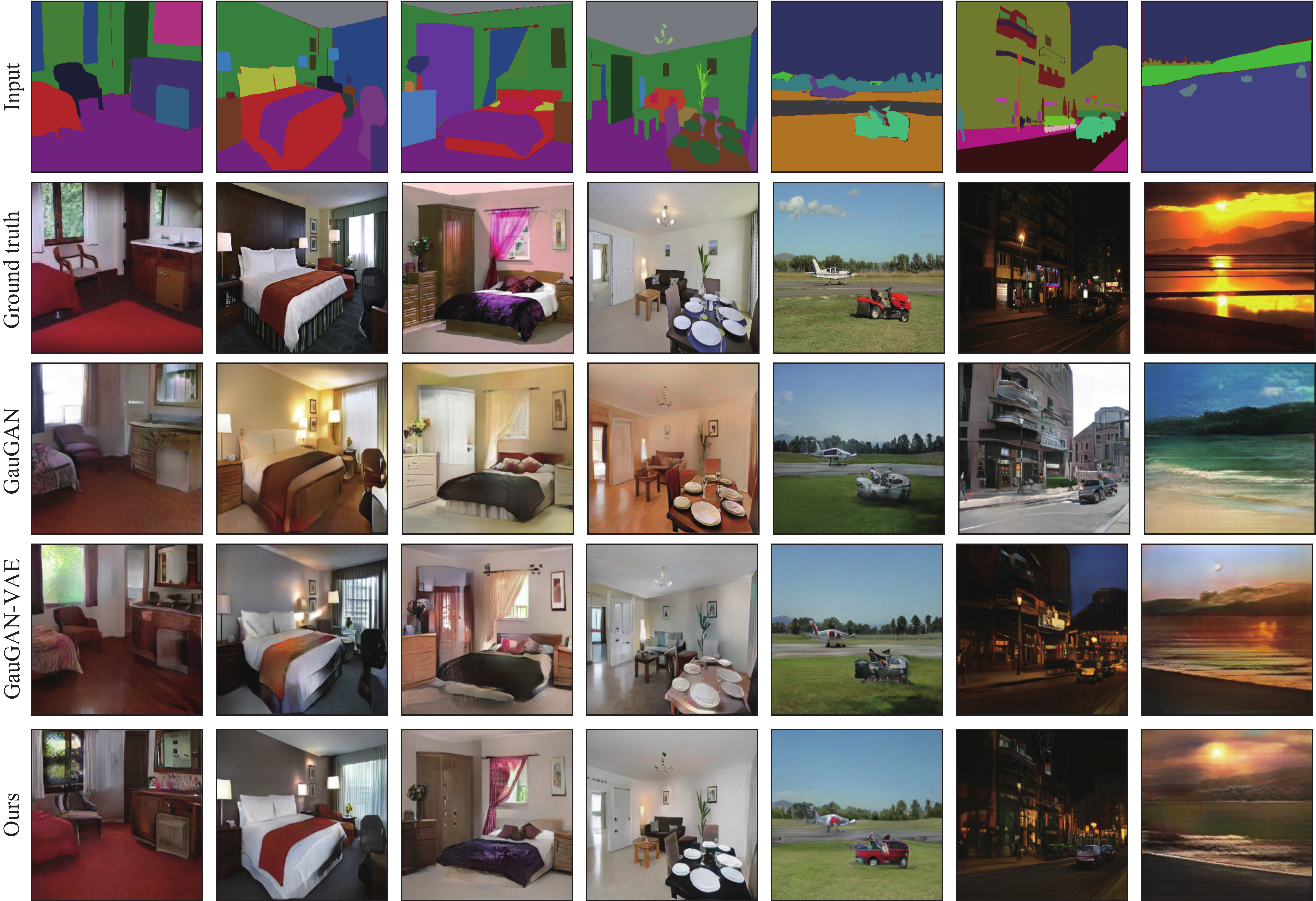}
   \caption{
Qualitative comparison of reconstruction performance using latent codes \revA{extracted from} the ground-truth images in the ADE20K dataset. 
Our method can reproduce styles for each object in \revC{the} ground-truth images more faithfully than the baseline \revC{methods}. 
\revG{To see the differences more clearly, please zoom in on the electronic version of \revH{this} paper.} 
   }
   \label{fig:results_1}
\end{figure*}

\section{Experiments}\label{sec:experiments}
\subsection{Datasets}

We \revA{used the following} three datasets \revA{in our experiments}. 
\begin{itemize}
 \item \revC{The} \textit{ADE20K}~\cite{DBLP:conf/cvpr/ZhouZPFB017} dataset contains images of indoor and outdoor scenes labeled with 151 semantic classes \revD{(including the ``unknown'' class)}, such as window, bed, sky, and tree. It consists of 20,210 training images and 2,000 validation images. 
 In the training set, we resized the images to $286\times 286$ and applied random cropping of $256\times 256$ for each iteration. 
 In the validation set, we resized the images to $256\times 256$. 
 \item \textit{DeepFashion}~\cite{liuLQWTcvpr16DeepFashion} (In-shop Clothes Retrieval Benchmark) is a fashion-image dataset \revA{with} 16 semantic classes (\revC{i.e.,} top, pants, hair, etc.). 
 We excluded \revC{the} photographs \revA{that do not contain face labels}.
 \revA{Our} training and test sets \revC{contained} 6,343 and 6,390 images, respectively. 
 We \revA{squared the images by padding them with} white rectangles \revA{at the left and right sides} and then resized them to $256 \times 256$. 
 \revA{Note that we removed these paddings from \revC{the} resultant images for \revJ{better visual} \revC{quality}.}
  \item \revC{The} \textit{GTA-5}~\cite{DBLP:conf/eccv/RichterVRK16} dataset includes rendered images of street scenes. Although \revC{Cityscapes is} a popular street photograph dataset~\cite{DBLP:conf/cvpr/CordtsORREBFRS16}, GTA-5 has much greater variations in terms of weather conditions and object appearance~\cite{DBLP:conf/iccv/LiZM19}. 
  We used the post-processed version of the GTA-5 dataset published \revC{in} \revA{the} recent work \revC{on} multimodal semantic image synthesis~\cite{DBLP:conf/iccv/LiZM19}. 
  This dataset contains 12,403 training images and 6,382 test images \revA{with} 20 semantic classes \revF{(19 classes that are common in GTA-5 and Cityscapes, plus an ``others'' class)},
  including sky, road, \revC{and} building. The image size is $512\times 256$. 
\end{itemize}

\subsection{Experimental Settings}

We implemented our method with PyTorch and conducted experiments on GeForce GTX 1080 Ti and Quadro RTX 6000. 
We used the same training parameters as in the GauGAN~\cite{DBLP:conf/cvpr/Park0WZ19} and GroupDNet papers~\cite{conf/cvpr/zhu20} \revJ{except for} the batch size. 
For optimization, we used Adam with the momentum term $\beta = (0.0, 0.9)$ and set the learning rates of the generator and encoder to 0.0004 and the discriminator to 0.0001. 
Due to our limited GPU resources, we set the batch size to 4 \revC{(the GauGAN paper uses a larger size)}. 
For the dimensions of the latent codes, we set $N_1=256$ and $N=C$ (where $C$ is the number of classes in the dataset).
In the ADE20K and DeepFashion datasets, we performed 100 and 60 \revD{epochs of} \revC{training}\revC{. These are followed by a} further 100 and 40 epochs \revA{with linear decays to zero for} the learning rate. 
In the GTA-5 dataset, 20 epochs of training were sufficient to learn plausible image translation. 
\revC{In addition,} we applied random horizontal flipping to the images for all datasets \revC{during training}. 

\subsubsection{Compared methods}
We compared our method with \revC{the} GauGAN~\cite{DBLP:conf/cvpr/Park0WZ19}, GauGAN-VAE~\cite{DBLP:conf/cvpr/Park0WZ19}, conditional implicit maximum likelihood estimation (cIMLE)~\cite{DBLP:conf/iccv/LiZM19}, and concurrent GroupDNet \revC{approaches}~\cite{conf/cvpr/zhu20}. 
For \revD{GauGAN}, we generated results using the model \revC{(published by the authors)} pre-trained on ADE20K.
For \revD{GauGAN-VAE}, we performed \revC{the} training with the same parameter \revC{settings} as our method because their pre-trained models are not publicly available. 
For \revC{the cIMLE model}, we generated results using the model pre-trained on GTA-5. 
To train our model and \revD{GauGAN-VAE} using the GTA-5 dataset, we adopt the rebalancing schemes for the dataset and the loss function \revA{used in cIMLE}, for fair comparison. 
For GroupDNet, we used the pre-trained model for the ADE20K dataset. 
\revC{However}, we trained their model from scratch for the DeepFashion dataset because their pre-processed test set and segmentation masks are not \revC{publicly available}. 
\revC{The} \revJ{inputs for our results} are not included in the training \revJ{datasets}. 

\subsection{Results and Analysis}
In this section, we discuss the experimental results to answer \revA{two research questions (RQs) as follows}.

\subsubsection*{RQ1. Can our model capture the style of each class using latent codes better?}
If \revA{our} model \revA{can} \revA{\revC{adequately}} capture the \revA{class-wise} styles, we should be able to reproduce the styles of a ground-truth image faithfully using \revC{the} appropriate latent codes. 
To evaluate this, we used the trained encoder as a style guidance network \revA{to extract \revC{the} ``ground-truth'' latent codes from \revC{the} ground-truth images}. 
We \revA{attempted to reproduce the ground-truth images in the test set by extracting the target styles from the images.}
\revC{The} generated images $G(E(\mathbf{M}, \mathbf{I}))$ are compared with the ground-truth images $\mathbf{I}$. 
As an evaluation metric, we used \revC{the} \textit{Fr\'{e}chet \revD{i}nception \revD{d}istance} (FID), 
which measures the distance between two image sets generated from a GAN distribution and a true distribution. Furthermore, to measure the similarity between the generated and ground-truth images one by one, we used \revA{the} \revC{\textit{learned perceptual image patch similarity}} (LPIPS)~\cite{DBLP:conf/cvpr/ZhangIESW18} and \revA{computed} the average \revC{for all images in the test set} (\revA{\revC{hereinafter} called \revC{the}} \revC{\textit{GT-similarity}}). 
\begin{figure*}[t]
   \centering
   \includegraphics*[width=1.\linewidth, clip]{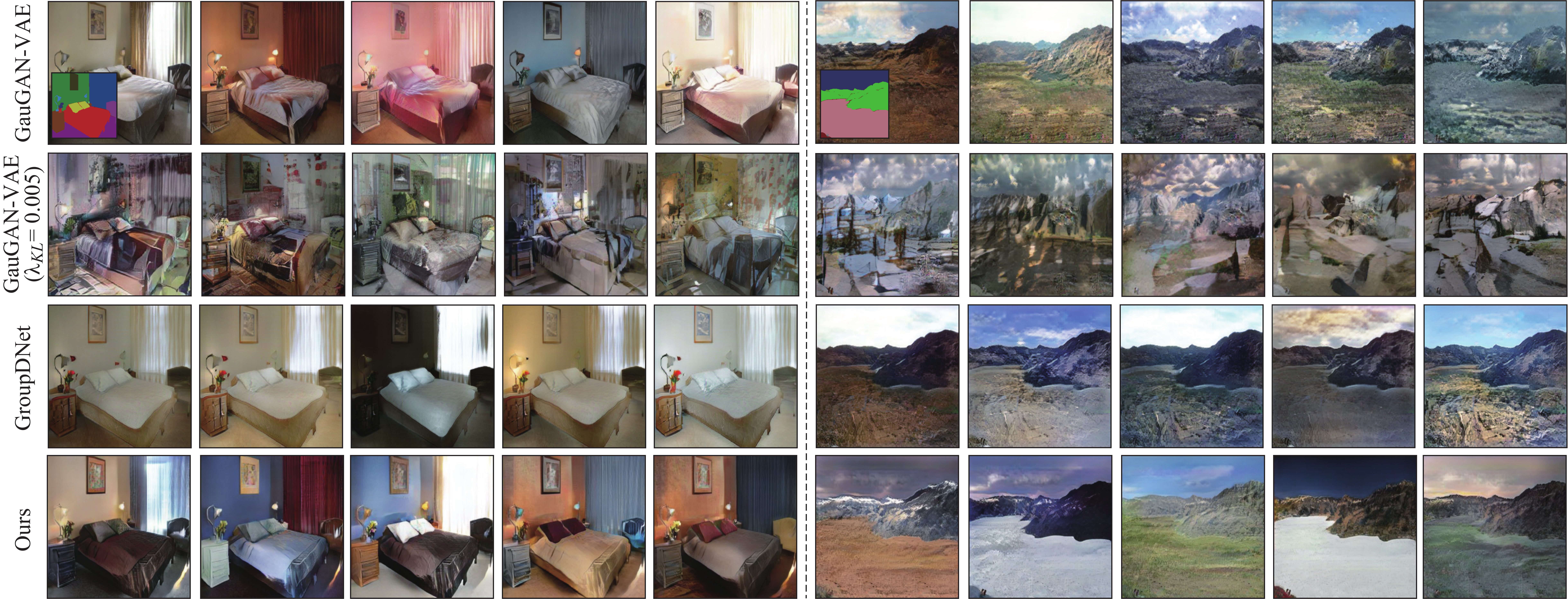}
   \caption{
Comparison of the results generated using \revC{the} \revA{randomly-sampled} latent codes \revA{with} the ADE20K dataset. 
The inset at \revA{each} upper left \revC{region} is the input semantic mask.
Our method can produce more diverse images than the baselines, without quality degradation. 
In \revD{GauGAN-VAE}, the weak regularization ($\lambda_{KL}=0.0005$) \revA{for  latent codes} increases the diversity of the generated images\revC{. However, it produces} more artifacts due to \revJ{the poor} generalization performance. 
   }
   \label{fig:results_2}
\end{figure*}

\begin{figure*}[t]
   \centering
   \includegraphics*[width=1.\linewidth, clip]{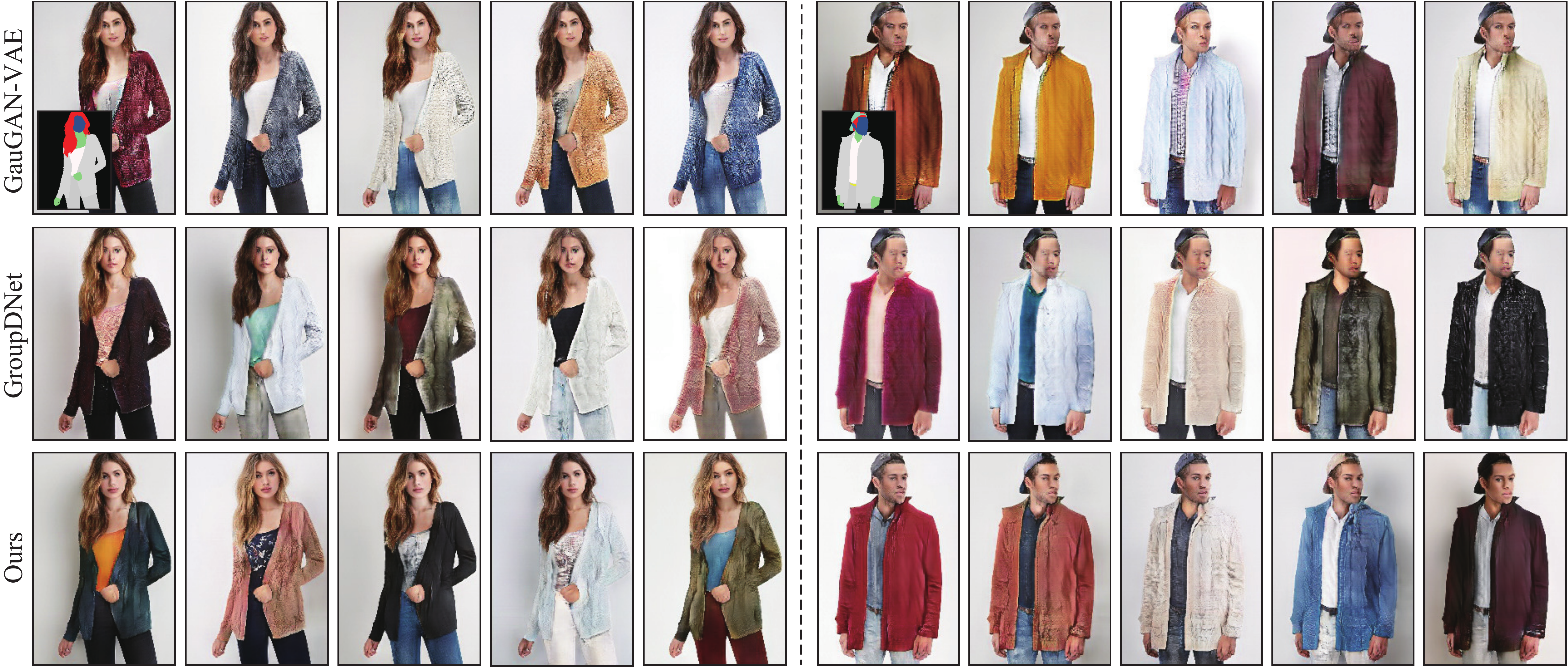}
   \caption{
Comparison of the results generated using \revC{the} \revA{randomly-sampled} latent codes \revA{with} the DeepFashion dataset.
   }
   \label{fig:results_2_2}
\end{figure*}

\begin{figure*}[t]
   \centering
   \includegraphics*[width=1.\linewidth, clip]{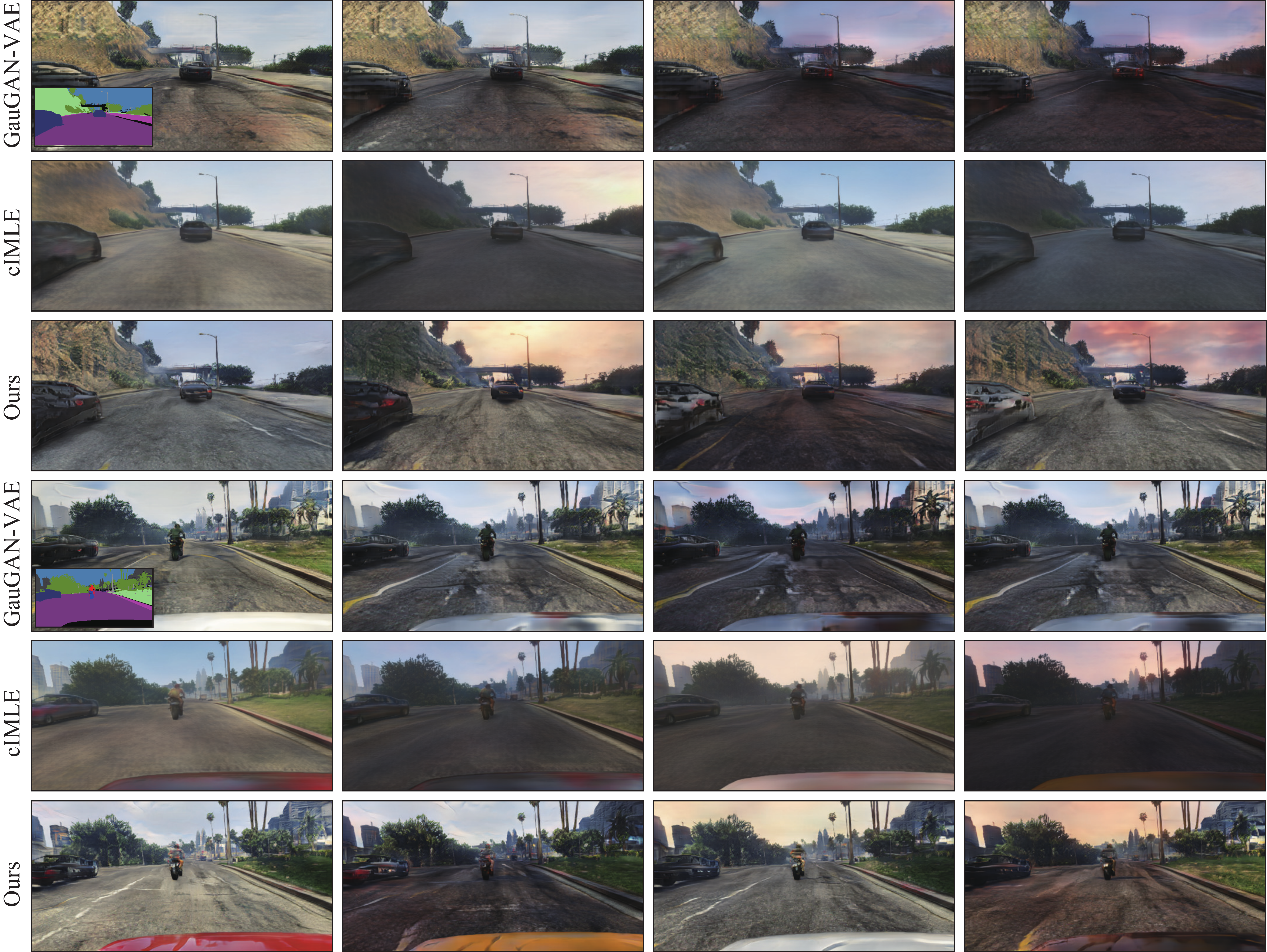}
   \caption{
Comparison of the results generated using \revC{the} \revA{randomly-sampled} latent codes \revA{with} the GTA-5 dataset.
    }
   \label{fig:results_2_3}
\end{figure*}

\begin{table}[t]
\centering
\caption{
Quantitative comparison of reconstruction performance using \revC{the} latent codes \revA{extracted from} \revC{the} ground-truth images in the ADE20K dataset. 
The \textcolor{red}{\textbf{red}} and \textcolor{blue}{\textbf{blue}} font \revJ{colors} denote the best and second-best scores, respectively. 
}
  \begin{tabular}{l||c|c} \hline
    Methods & FID $\downarrow$ & \revC{GT-similarity}$\downarrow$ \\ \hline \hline
    GauGAN & 33.9 & 0.538 \\
    GauGAN-VAE & \textcolor{blue}{\textbf{32.9}} & \textcolor{blue}{\textbf{0.493}}  \\
    Ours & \textcolor{red}{\textbf{31.2}} & \textcolor{red}{\textbf{0.462}} \\ \hline
  \end{tabular}
\label{table:score1}
\end{table}

Table~\ref{table:score1} shows a quantitative comparison \revC{of various methods} in the ADE20K dataset\revC{; the} results show that our method \revA{yields} better values in the two metrics. 
Because \revD{GauGAN}~\cite{DBLP:conf/cvpr/Park0WZ19} \revA{does not accept user-specified} styles, \revC{the GT-similarity score \revJ{degrades} even if the FID score is somewhat satisfactory.}
On the other hand, our method shows greater improvement in \revC{GT-similarity} compared to \revD{GauGAN-VAE}
~\cite{DBLP:conf/cvpr/Park0WZ19}. 
These results \revA{indicate} that our model \revA{can \revC{better} capture} the style of each class using latent codes.
In the qualitative comparison shown in Figure~\ref{fig:results_1}, our results are visually more faithful to the ground-truth images than \revC{those of the compared methods.} 
Specifically, individual objects \revC{have} similar styles to the ground-truth \revJ{images}\revC{; for example}, floor, wall, and bed \revC{images} in the indoor scenes, and car, sky, and sunset in the outdoor scenes. 
\revA{Although \revC{the} ground-truth images are not available in actual applications}, our method can generate plausible images with specific styles if we have other reference images containing common classes (see Section~\ref{sec:style}).

\begin{table}[t]
\centering
\caption{
Quantitative comparison of the results generated using \revC{the} \revA{randomly-sampled} latent codes \revA{with} the ADE20K dataset. 
GauGAN-VAE* means GauGAN-VAE with weakened regularization ($\lambda_{KL} = 0.0005$) \revA{for latent codes}.
}
  \begin{tabular}{l||c|c} \hline
    Methods & Diversity  $\uparrow$ & Top 1 \revC{GT-similarity} $\downarrow$\\ \hline \hline
    GauGAN-VAE & 0.294 & \textcolor{red}{\textbf{0.490}}\\
    GauGAN-VAE* & \textcolor{red}{\textbf{0.507}} & 0.583 \\
    GroupDNet & 0.177 & 0.521 \\
    Ours & \textcolor{blue}{\textbf{0.415}} & \textcolor{blue}{\textbf{0.503}}\\ \hline
  \end{tabular}
\label{table:score2}
\end{table}

\begin{table}[t]
\centering
\caption{
 Quantitative comparison of the results generated using \revC{the} \revA{randomly-sampled} latent codes \revA{with} the DeepFashion dataset.
 }
  \begin{tabular}{l||c|c} \hline
    Methods & Diversity $\uparrow$ & Top 1 \revC{GT-similarity} $\downarrow$ \\ \hline \hline
    GauGAN-VAE & 0.203 &  \textcolor{red}{\textbf{0.155}}\\
    GroupDNet & \textcolor{red}{\textbf{0.218}}& 0.159\\
    Ours & \textcolor{blue}{\textbf{0.216}} & \textcolor{blue}{\textbf{0.158}}\\ \hline
  \end{tabular}
\label{table:score3}
\end{table}

\subsubsection*{RQ2. Can our model generate diverse and perceptually plausible images?}
We \revA{evaluated \revD{the}} results generated using \revC{the} latent codes \revA{randomly} sampled from the prior without specifying specific style \revA{images}. 
For diversity, we used the LPIPS score\revA{,} following \revA{the work \revC{of}} Li et \revJ{al.}~\cite{DBLP:conf/iccv/LiZM19}.
We generated 40 images for each test image using \revA{randomly sampled} latent codes. 
\revC{Next, we} computed the LPIPS values for each pair in 40 images and took the average of them.
By calculating \revD{this for all the test images and averaging them,}
we obtained the \textit{Diversity} score\revA{; higher scores indicate greater diversity}. 
As for visual plausibility, we can evaluate some results \revA{subjectively} via qualitative comparison. 
However, objective evaluation is challenging because \revC{not all} the randomly generated images have corresponding ground-truth images. 
To deal with this, we selected the image \revJ{closest} to the ground-truth image \revC{regarding the} $L_1$ distance among the generated images\revC{, also following the work of Li et al.}. 
For all test images, we computed \revA{the} LPIPS scores between the selected images and the ground-truth images\revA{,} and used the average of those top 1 similarities as an objective plausibility score (\revA{\revC{hereinafter} called \revC{the}} \textit{Top 1 \revC{GT-similarity}}). 

\begin{figure}[t]
   \centering
   \includegraphics*[width=1.\linewidth, clip]{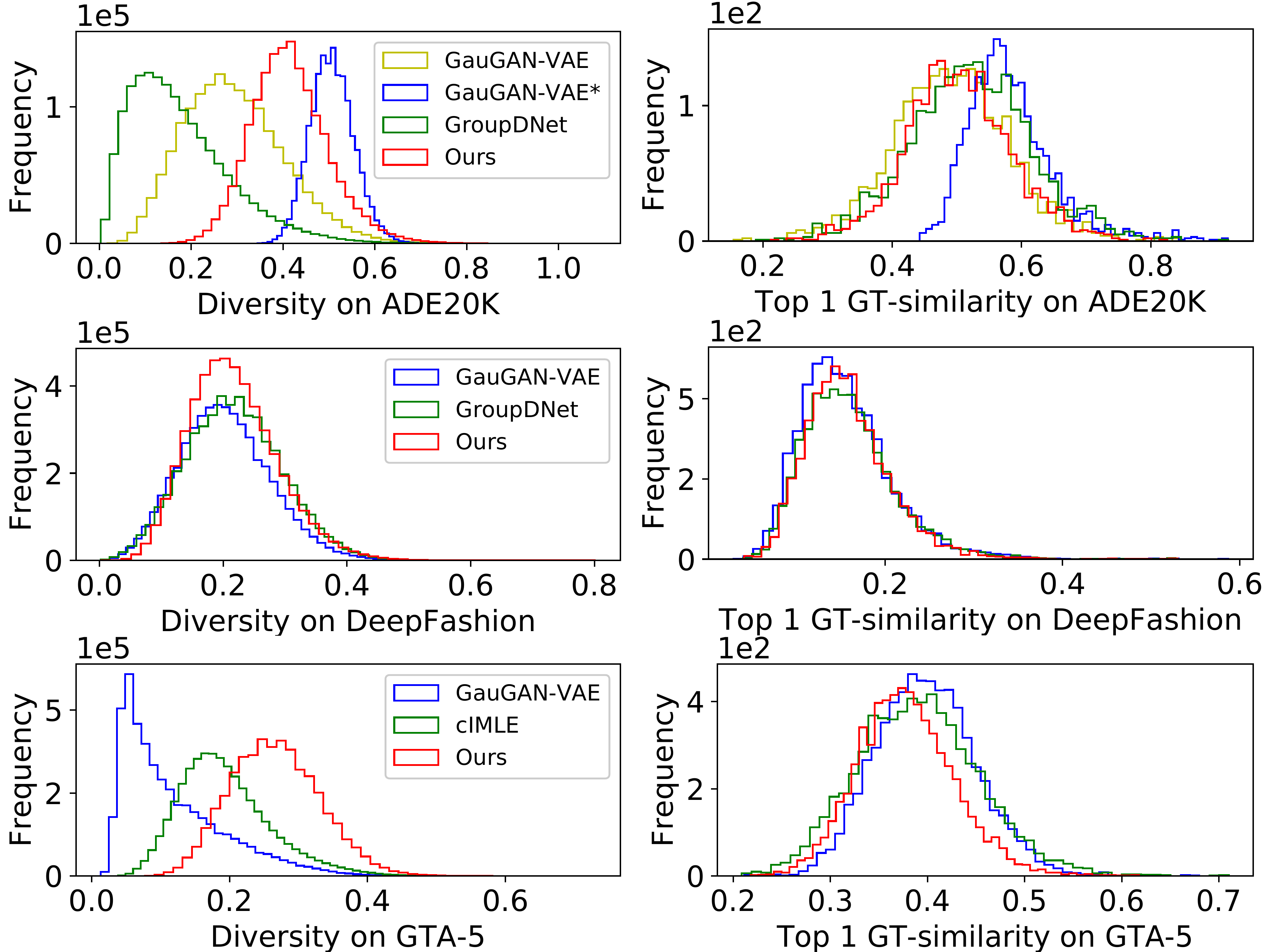}
   \caption{
\revG{Histograms of Diversity (larger is better) and Top 1 GT-similarity (smaller is better) on each dataset. }
    }
   \label{fig:histogram}
\end{figure}

\begin{table}[t]
\centering
\caption{
Quantitative comparison of the results generated using \revC{the} \revA{randomly-sampled} latent codes \revA{with} the GTA-5 dataset. 
}
  \begin{tabular}{l||c|c} \hline
    Methods & Diversity $\uparrow$ & Top 1 \revC{GT-similarity} $\downarrow$ \\ \hline \hline
    GauGAN-VAE & 0.118 & 0.399\\
    cIMLE &  \textcolor{blue}{\textbf{0.194}} & \textcolor{blue}{\textbf{0.390}} \\
    Ours & \textcolor{red}{\textbf{0.266}} & \textcolor{red}{\textbf{0.378}} \\ \hline
  \end{tabular}
\label{table:score4}
\end{table}

\begin{figure}[t]
   \centering
   \includegraphics*[width=0.6\linewidth, clip]{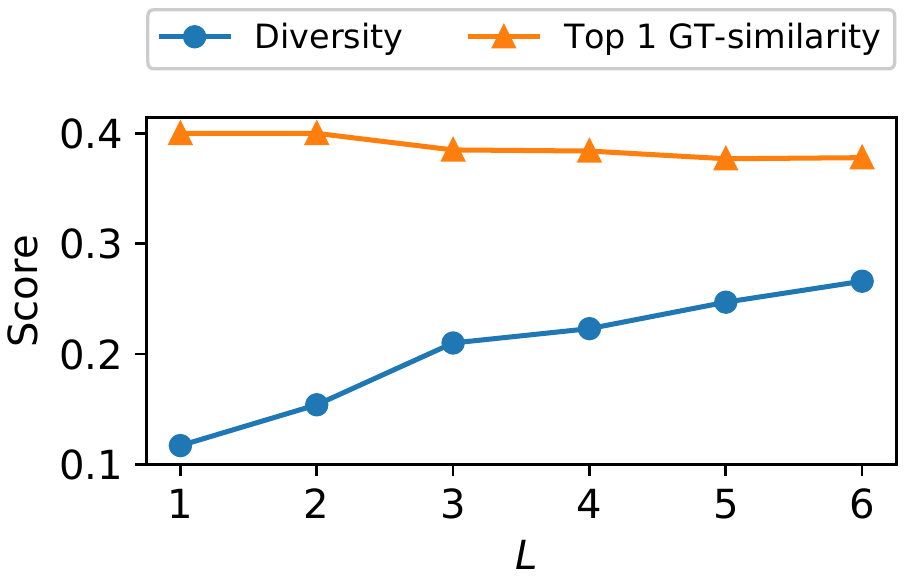}
   \caption{
\revG{Evaluation scores \revH{with different numbers} of encoder layers $L$ on the GTA-5 dataset.}
    }
   \label{fig:n_layers}
\end{figure}

\begin{table}[t]
\centering
\caption{
\revG{Ablation study on the GTA-5 dataset. }
}
  \begin{tabular}{l||c|c} \hline
    Methods & Diversity $\uparrow$ & Top 1 \revC{GT-similarity} $\downarrow$ \\ \hline \hline
    W/o class-wise VAE & 0.131 & 0.390\\
    W/o layer-wise VAE & 0.242 & 0.378\\
    Ours-50\% & 0.210 & 0.385\\
    Ours-75\% & 0.247 & 0.377 \\
    Ours& 0.266 & 0.378 \\ \hline
  \end{tabular}
\label{table:ablation}
\end{table}

\begin{figure}[t]
   \centering
   \includegraphics*[width=1.\linewidth, clip]{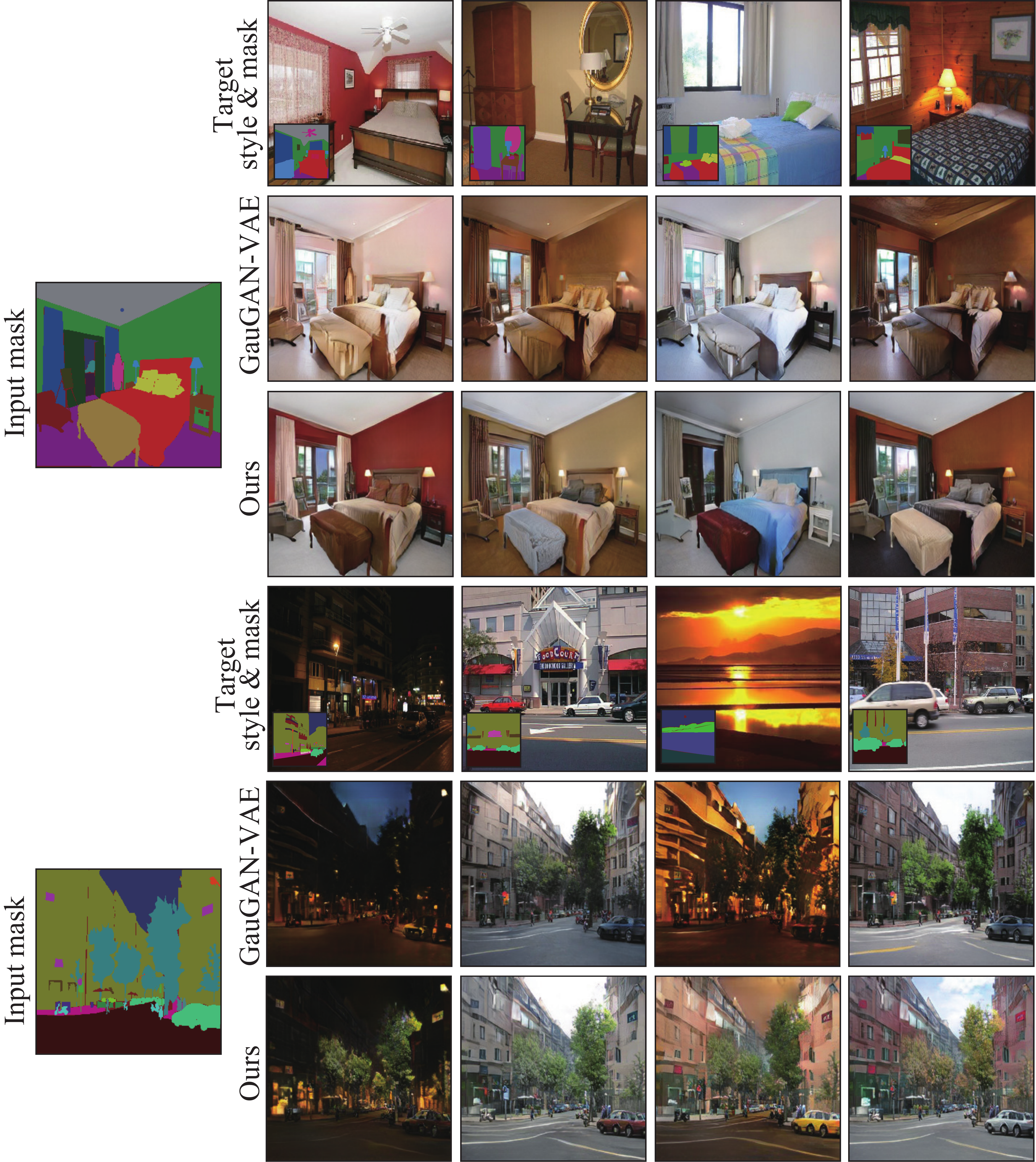}
   \caption{
Style guided synthesis in the ADE20K dataset. 
Our method can synthesize \revA{various images according to} the target styles of each object using \revC{the} \revA{target} semantic \revA{masks (insets)}. 
Our results are more faithful to the target styles than the baseline \revC{methods}.
}
   \label{fig:style}
\end{figure}

Table~\ref{table:score2} shows quantitative results for the ADE20K dataset. 
The Diversity \revA{scores} indicate that our method can generate more diverse images than \revC{both} GauGAN-VAE~\cite{DBLP:conf/cvpr/Park0WZ19} and GroupDNet~\cite{conf/cvpr/zhu20}. 
\revC{Conversely}, there is no \revC{significant} difference in Top 1 \revC{GT-similarity}. 
We also tested GauGAN-VAE with weakened regularization of $\mathcal{L}_{KL}$ \revA{to make the results \revC{comparatively} more diverse}. 
Although the diversity was improved, the generated image \revJ{degraded} severely, as shown in the Top 1 \revC{GT-similarity} score and qualitative results in Figure~\ref{fig:results_2}.
This \revC{degradation is because of the} large discrepancy between the prior and the distribution \revA{that} the trained generator assumes. 
Furthermore, \revA{\revC{though}} GauGAN-VAE \revC{globally} changes the appearance of the images,  
our method reproduces local and class-wise variations while retaining image quality. 
For the DeepFashion dataset (Table~\ref{table:score3} and Figure~\ref{fig:results_2_2}), there is no significant difference between each method in terms of image quality. 
As for diversity, \revC{both} GroupDNet and \revC{our model} are modestly better than GauGAN-VAE. 
\revD{GauGAN-VAE} can reproduce diverse appearances in large classes\revC{,} such as jackets\revC{,} but cannot handle small classes such as pants, inner shirts, and hair. 
GroupDNet \revJ{obtained} diverse appearances for each item,
\revJ{but} it \revA{suffered from a large} number of classes as demonstrated in the ADE20K results. 
\revC{By} contrast, our method works \revC{well in} both cases.  
In Table~\ref{table:score4} and Figure~\ref{fig:results_2_3}, we also \revC{ran comparisons} with cIMLE~\cite{DBLP:conf/iccv/LiZM19} using the GTA-5 dataset.
The results demonstrate that cIMLE can generate more diverse images than GauGAN-VAE. 
However, as shown in the qualitative results, cIMLE handles global color variations only. 
The results \revC{for} cIMLE are also smooth, while the GAN-based methods give finer outputs. 
\revC{By} contrast, our method \revC{showed} good performance \revC{both} quantitatively and qualitatively. 
\revH{To visualize the detailed statistical trends, we further show the histograms of evaluation scores in Figure~\ref{fig:histogram}. We can see that each distribution is unimodal, and our results are consistently better or comparable on average.}
We \revC{provide} more samples for these results in the supplementary materials.

\subsubsection{Ablation study}
We also conducted an ablation study to validate the effectiveness of \revH{our} class-wise VAE and layer-wise VAE, respectively. 
Table~\ref{table:ablation} shows the evaluation scores of our method with and without our VAE modules. 
\textit{W/o class-wise VAE} means that a single latent code is sampled from each layer,
whereas \textit{W/o layer-wise VAE} means that class-wise latent codes are sampled \revH{only from} the middle layer. 
The results demonstrate that both VAEs are capable of diversifying generated images without degrading their quality. 
\revH{Compared to \textit{Ours}, we can see that both of our VAE extensions help us diversify} generated images without degrading their quality.

In Figure~\ref{fig:n_layers}, we further investigated the effect of the number of encoder layers $L$, which is \revH{six} in \revH{\textit{Ours}}. 
\revH{For} $L<6$, we \revH{adjusted the sizes of} the feature maps before the final \revH{linear} layer to the \revH{size of} $L=6$. 
 \textit{Ours-50\%} and \textit{Ours-75\%} in Table~\ref{table:ablation} mean that $50\%$ ($L=3$) and $75\%$ ($L=5$) of the encoder layers \revH{were} used. 
\revI{Generally, diversifying generated images often causes degradation of the image quality due to overfitting (like GauGAN-VAE* in Table~\ref{table:score2}). Nonetheless, these results indicate that increasing the number of layers makes generated images more diverse (i.e., higher Diversity) while keeping their quality (i.e., almost the same but a bit lower Top 1 GT-similarity). The computational cost increases approximately linearly according to the number of layers. For example, the training times for $L=1$, $2$, and $3$ were $46$, $55$, and $66$ minutes for one epoch. }

\subsection{Applications}\label{sec:apps}
In this section, we introduce three \revA{applications of} image synthesis and editing based on our method. 

\begin{figure}[t]
   \centering
   \includegraphics*[width=1.\linewidth, clip]{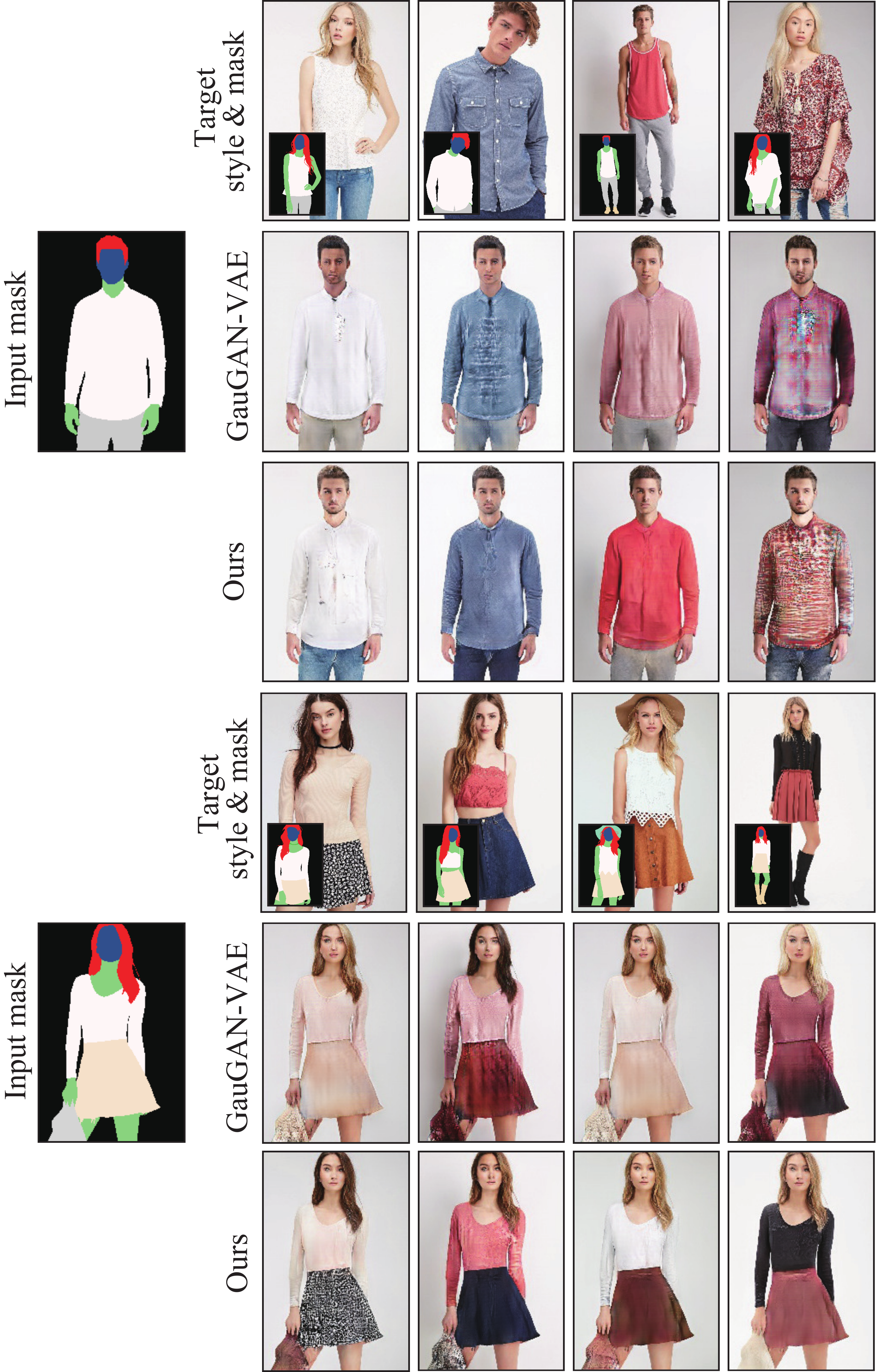}
   \caption{
Style guided synthesis in the DeepFashion dataset.
}
   \label{fig:style_2}
\end{figure}

\begin{figure}[t]
   \centering
   \includegraphics*[width=1.\linewidth, clip]{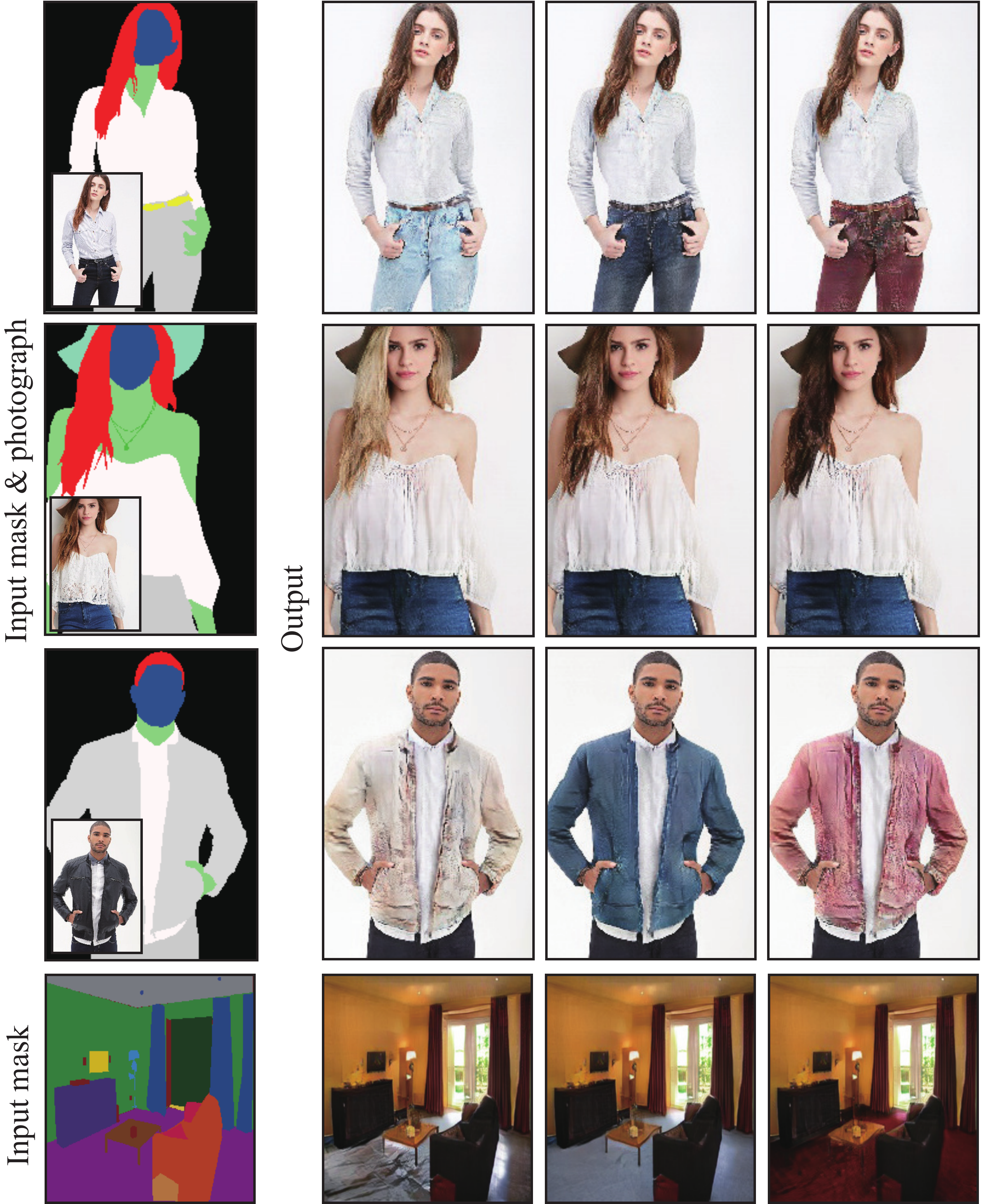}
   \caption{
Object editing. 
In the top three rows, \revA{we changed \revC{the} regions of specific classes\revC{,} such as clothes and hair using randomly-sampled latent codes} while preserving \revA{\revC{the} other regions}. 
In the bottom row, \revA{we} generated the image from the semantic mask and edited the floor design using \revC{the} latent codes. 
}
   \label{fig:results_3}
\end{figure}

\begin{figure}[t]
   \centering
   \includegraphics*[width=1.\linewidth, clip]{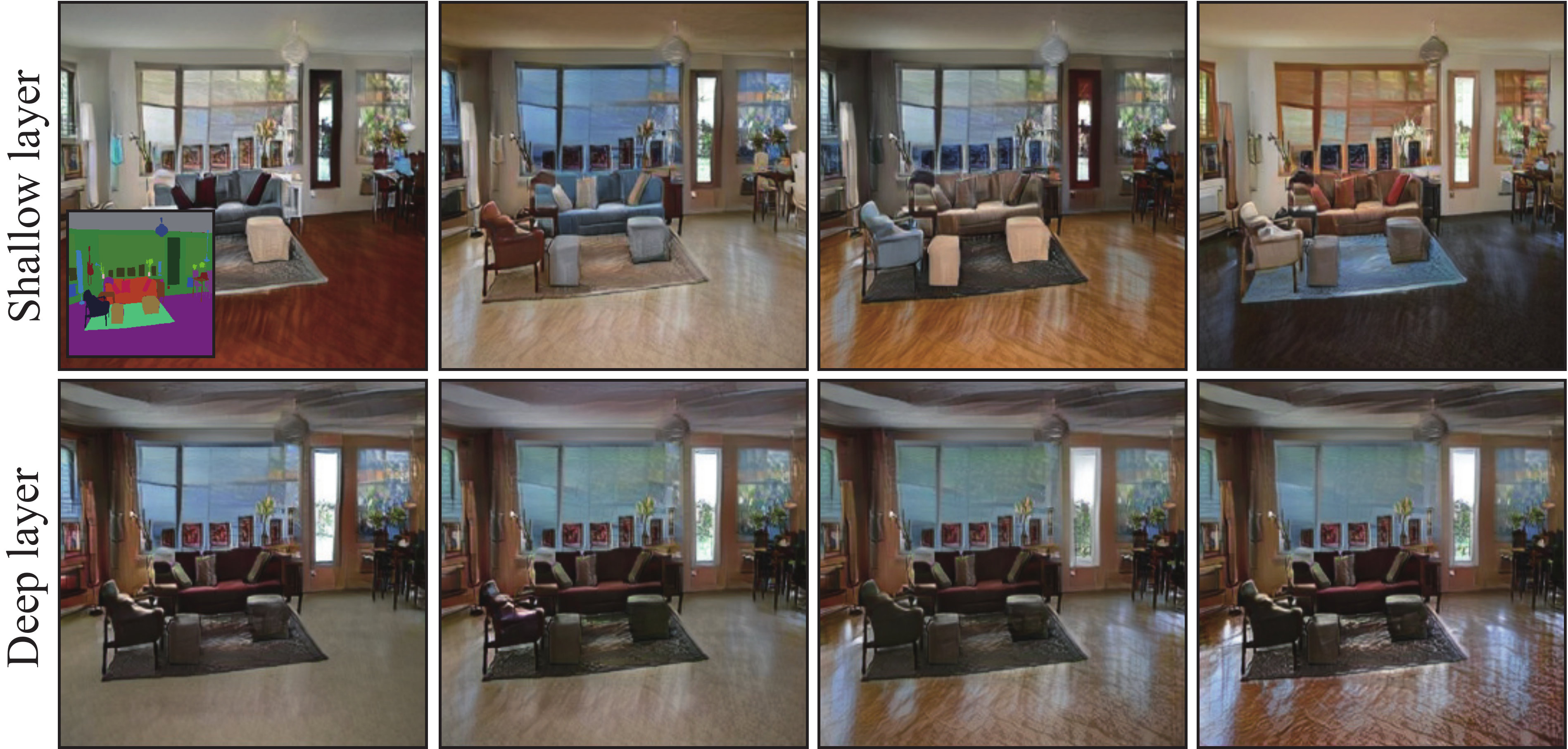}
   \caption{
Material editing. 
We can control \revA{the} global appearance (e.g., color of entire objects) by moving the latent codes \revA{at} the shallow layer \revA{in} the generator\revC{. Likewise,} we can control \revA{the} local appearance (e.g., edges and textures) by moving the latent codes \revA{at} the deeper layers. 
}
   \label{fig:results_4}
\end{figure}

\subsubsection{Style guided synthesis}\label{sec:style}
\revA{From} a source mask $\mathbf{M}_s$, \revA{our} method can synthesize an image \revA{in reference to} the styles \revA{extracted from} an arbitrary target image $\mathbf{I}_t$ using the pre-trained encoder. 
\revA{Specifically, we take \revC{the} latent codes from $E_{l,c}(\mathbf{I}_t)$ for \revC{the} classes that appear both in the source \revJ{and} target masks $\mathbf{M}_s$ and $\mathbf{M}_t$\revC{;} i.e., $\mathcal{C}(\mathbf{M}_s) \cap \mathcal{C}(\mathbf{M}_t)$. \revC{In addition, we} take random samples from $\mathcal{N}(\mathbf{0},I)$ for \revC{the $\mathbf{M}_s$} classes that are not included in $\mathbf{M}_t$\revC{;} i.e., $\mathcal{C}(\mathbf{M}_s) \cap \overline{\mathcal{C}(\mathbf{M}_t)}$.}
The results are shown in Figures~\ref{fig:style} and~\ref{fig:style_2}. 
Our method reproduces the styles of the individual objects such as walls, sheets, clothes, and hair more faithfully than GauGAN-VAE.
\revA{This application is similar to} \revD{image analogies}~\cite{DBLP:conf/siggraph/HertzmannJOCS01}, 
which applies a transformation between a mask $A$ and an image $A'$ to a mask $B$. 
Our method differs from this technique in that classes included in the mask $B$ \revC{are} not necessarily included in the mask $A$. 
\revD{Our} method can handle missing classes via latent codes \revJ{sampled} from the prior. 

\subsubsection{Object editing}

As \revA{a design exploration tool}, 
our method is useful for object appearance editing by manipulating \revC{the} latent codes for a particular class. 
\revC{The} bottom of Figure~\ref{fig:teaser} and the top three rows of Figure~\ref{fig:results_3} \revC{show}
\revC{the appearances} of pants, hair, etc. \revJ{edited} \revA{individually while retaining the styles of \revC{the} other classes.} 
To maintain \revC{a} person's identity, we preserved the original faces and skin \revA{by copying the pixel values from the input photographs}. 
\revC{The} bottom of Figure~\ref{fig:results_3} \revC{shows \revJ{that} the} different latent codes generate \revC{the} various floor patterns, \revC{such as} marble or wood\revC{,} with different glossiness. 
A recent study by Bau et \revA{al.}~\cite{DBLP:journals/tog/BauSPWZZ019} also \revC{attempted} a similar task. 
Their method can synthesize photorealistic images by optimizing latent representation based on \revC{the} image prior learned by \revC{the} GANs. 
While this optimization requires tens of seconds, our method does not require such optimization. 

\subsubsection{Material editing}
We can \revA{also} edit \revA{object} materials by manipulating \revC{the} \revA{layer-wise} latent codes. 
In Figure~\ref{fig:results_4}, \revC{it is evident that} moving the latent codes fed to the shallow layer changes \revC{the} global appearance, \revA{e.g.,} the color of an entire object. 
\revC{Conversely}, moving \revC{the} latent codes fed to the deep layer changes \revC{the} local appearance \revC{features}, such as edges and textures. 
However, the latent codes for some layers did not have \revC{as} much effect on \revC{the} appearance variations as we expected. 
\revC{The most likely reason for this is that} the degree of \revC{the} freedom of latent codes \revC{was} too large. 
In \revA{future work}, we \revA{would like to determine} \revC{the} appropriate dimensions of \revC{the} latent codes \revA{automatically} according to \revC{the} data variation in a target domain. 

\subsection{Limitations}
\begin{figure}[t]
   \centering
   \includegraphics*[width=1.\linewidth, clip]{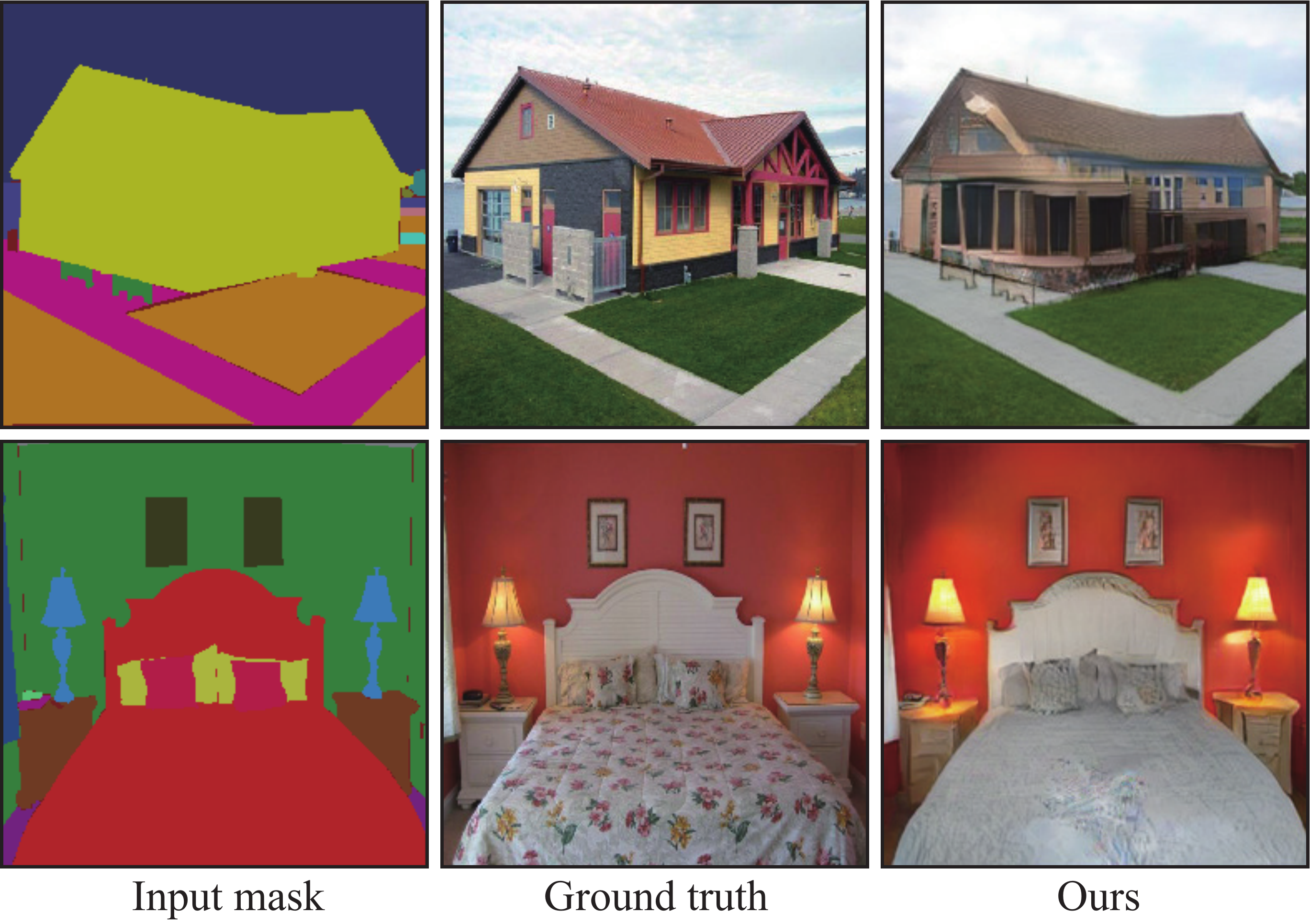}
   \caption{
   Our failure cases. 
   \revA{\revC{Faithfully reproducing}} complex structures and textures \revA{is difficult} even \revJ{with} \revC{the} latent codes extracted from the ground-truth images.
   }
   \label{fig:limitations}
\end{figure}

Although our method \revC{significantly} improves the diversity of synthesized images while maintaining generalization performance, some limitations remain. 
First, there is room for improvement in style reproducibility. 
Figure~\ref{fig:limitations} shows the \revC{images} reconstructed using \revC{the} latent codes extracted from the ground-truth images. 
The results show that complex patterns, such as the window position and the bed texture, are not completely reproduced.
Simply increasing the \revC{dimensions} of \revC{the} latent codes would improve the expressiveness of the model but could worsen the generalization performance due to overfitting. 
To \revC{solve} this problem, we \revD{seek} a more efficient disentanglement representation in future \revC{work}. 
Nevertheless, our \revC{method makes a significant contribution by diversifying} semantic image synthesis while avoiding \revC{output} degradation. 

Another limitation \revC{of our method} arises from \revA{our assumption that} each latent code is independent for each class and layer. 
This assumption \revC{can} sometimes \revA{\revC{cause} context mismatch in \revC{the} object appearance}. 
For example, \revC{as shown} in the \revE{third} row and \revE{fourth} column in Figure~\ref{fig:results_2_3}, 
\revD{it} \revC{is possible that} \revE{users} might \revA{expect} the road \revC{to} be \revC{redder in color} due to \revC{the effect of} the sunset. 
Although our method \revA{recognizes} scene contexts using the large receptive fields of the decoder,
\revC{these} results might not be sufficient. 
\revD{\revE{Users} can optionally repeat re-sampling of latent codes until he/she is satisfied, but we would like to develop a better approach that exploits the relationship among latent codes.}

\section{Conclusions}
In this \revC{study}, we have proposed a method for semantic image synthesis and editing via class- and layer-wise \revE{VAEs} \revA{to diversify output images}. 
Our network design allows for the acquisition of latent codes depending on semantic classes and layers. 
\revC{Furthermore, our} method is \revD{a} simple yet effective \revC{approach to improve} the expressivity of the model and \revC{capture} diverse styles by learning multiple latent spaces. 
\revA{Through experiments with three challenging datasets}, we have \revA{demonstrated} that our method can synthesize \revC{images that are both plausible and} diverse from a simple semantic mask. 
We have also \revA{shown} applications \revA{of our method} for style guided synthesis, object editing, and material editing. 
Future work \revJ{includes} \revC{the} adaptive adjustment of the appropriate degrees of freedom of \revC{the} latent codes. 
\revC{In addition, we} are interested in obtaining efficient disentanglement representation that \revC{considers} the correlation between classes.

\printbibliography

\end{document}